\pdfoutput=1
\documentclass[11pt]{article}

\usepackage[preprint]{acl}

\usepackage{times}
\usepackage{latexsym}
\usepackage{amsfonts}
\usepackage{subfigure}
\usepackage{graphicx}
\usepackage{amsmath} 
\usepackage{multirow}
\usepackage{xcolor}
\usepackage{multirow}
\usepackage{colortbl}

\usepackage[T1]{fontenc}
\usepackage[utf8]{inputenc}

\usepackage{microtype}

\usepackage{inconsolata}

\usepackage{graphicx}
\usepackage{tabularray}
\title{Interpreting token compositionality in LLMs: A robustness analysis}

\author{Nura Aljaafari$^{1\dagger}$,~ Danilo S. Carvalho$^{1,3}$,~ Andr\'{e} Freitas$^{1,2,3}$ \\
  $^{1}$ Department of Computer Science, University of Manchester, United Kingdom\\
  $^{2}$ Idiap Research Institute, Switzerland\\
  $^{3}$ National Biomarker Centre, CRUK-MI, Univ. of Manchester, United Kingdom\\
  \texttt{\{firstname.lastname\}@[postgrad.]$^{\dagger}$manchester.ac.uk}}

\begin{document}
\maketitle

\begin{abstract}
Understanding the internal mechanisms of large language models (LLMs) is integral to enhancing their reliability, interpretability, and inference processes. We present Constituent-Aware Pooling (CAP), a methodology designed to analyse how LLMs process compositional linguistic structures. Grounded in principles of compositionality, mechanistic interpretability, and information theory, CAP systematically intervenes in model activations through constituent-based pooling at various model levels. Our experiments on inverse definition modelling, hypernym and synonym prediction reveal critical insights into transformers' limitations in handling compositional abstractions. No specific layer integrates tokens into unified semantic representations based on their constituent parts. We observe fragmented information processing, which intensifies with model size, suggesting that larger models struggle more with these interventions and exhibit greater information dispersion. This fragmentation likely stems from transformers' training objectives and architectural design, preventing systematic and cohesive representations. Our findings highlight fundamental limitations in current transformer architectures regarding compositional semantics processing and model interpretability, underscoring the critical need for novel approaches in LLM design to address these challenges.
\end{abstract}

\section{Introduction}
Large language models (LLMs) based on Transformer architectures have rapidly expanded in scope and capability, demonstrating strong performance across a wide range of NLP tasks. However, critical limitations remain, including hallucinations, poor interpretability, and limited semantic transparency. One open challenge concerns \textit{linguistic compositionality}: how models combine smaller units of text (e.g., morphemes, words, phrases) into coherent meaning structures, and how this process is reflected in internal representations.

\begin{figure*}[ht]
    \centering
    \includegraphics[width=0.99\linewidth]{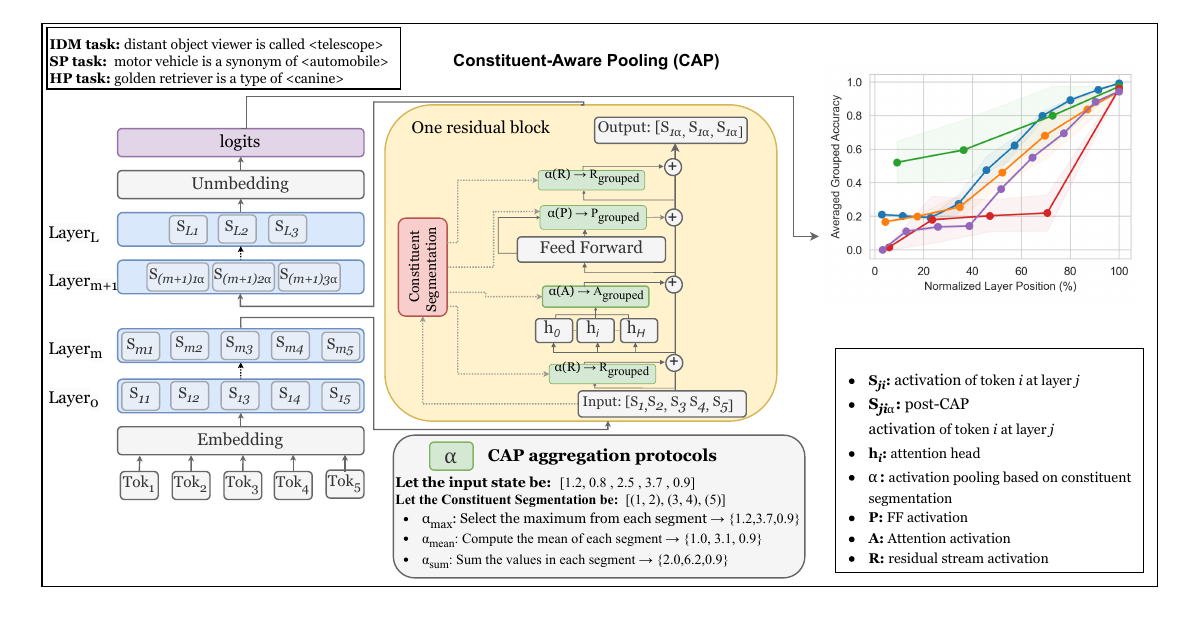}
    \caption{Illustration of the CAP process. Constituent segmentation identifies linguistic units (e.g., words or phrases), and CAP pools their activations at layer $m$ using aggregation (e.g., max, mean, sum). This operation reduces sequence length, and the modified activations are propagated to layer $m{+}1$. The results graph shows task accuracy under CAP at different depths.}
    \label{fig:cap_method}
\end{figure*}

Understanding how and where compositional structure is encoded in LLMs is essential for bridging the gap between user intent and model behaviour. Prior work has explored this by aligning model inputs and outputs~\cite{yin2024compositional}, embedding spaces~\cite{haslettmuch2024}, or layer-wise activations~\cite{yu2020assessing, modarressi2023decompx} with expected semantic representations. These approaches are grounded in two intuitive assumptions: (1) that LLMs internally represent compositional structure at the token or word level, and (2) that this information should be at least partially localisable at specific layers during inference.

Several studies have revealed that LLMs are often brittle under perturbation~\cite{wang2023large, fodor2024compositionality, hu2024prompt}, and that phrase-level representations may fail to align with expected semantics~\cite{carvalho2022montague}. Despite this, the mechanisms behind such fragility, particularly at the level of internal activations, remain poorly understood.

To investigate this, we propose \textit{Constituent-Aware Pooling (CAP)}, a structured perturbation method that groups token-level activations into larger constituent units (e.g., words or phrases) at arbitrary layers. CAP enables systematic probing of whether, and where, semantic meaning is robustly composed within the model. By applying CAP at varying depths, we assess the fragility of internal representations to compositional perturbations and examine whether, and how, semantic abstraction is distributed across layers.

Our empirical findings challenge common assumptions of hierarchical semantic buildup. Rather than gradually constructing compositional meaning across layers, LLMs often retain token-level focus well into the middle layers. Applying CAP, even at semantically coherent groupings, results in substantial accuracy degradation, especially in earlier layers. Surprisingly, larger models are more sensitive to such perturbations than smaller ones, suggesting increased representational fragility with scale.

We contextualise these results using an information-theoretic lens, proposing that Transformer models delay aggregation to maximise token-level information throughput. This leads to distributed, rather than localised, composition across layers, resulting in longer dependency paths and reduced mutual redundancy at each layer.

\vspace{0.5em}
\noindent In summary, our contributions are:
\begin{itemize}
    \item \textbf{A systematic analysis} of how current LLMs handle constituent-level composition, evaluated via CAP across layers, models, and tasks.
    \item \textbf{A theoretical explanation} grounded in information theory, suggesting that LLMs optimise for prediction by postponing semantic integration, thus fragmenting compositional meaning across depth.
\end{itemize}

We conclude that compositional semantics are not reliably localisable within any fixed layer of standard Transformer models. This holds across model scales, supervision types, and inference tasks, and instead appears tied to architectural depth. Our results suggest that recovering explicit compositional structure may require specialised training objectives or architectural constraints. Supporting code and datasets are available at a public repository\footnote{< anonymised url>}.

\section{Tokenisation and compositionality in LLMs}

Intuitively, aggregating the representations of tokens that compose a single meaning unit (e.g., averaging the embeddings of `m', `amm' and `al' to form a single token embedding) and then to larger phrasal units (e.g. adjectival and noun compositions), would have a relatively small impact on model inference, since they have a strong dependence on each other in a given context and thus share significant information. However, it has been shown that LLMs are highly sensitive to token placement \cite{yin2024compositional, hu2024prompt} and that their internal representations have no significant correlation with phrasal composition semantics~\cite{yu2020assessing, carvalho2022montague}.

The observed disconnection between LLM internal representations and linguistic knowledge regarding compositionality raises practical and theoretical questions towards the robustness of such models to perturbations strictly tied to compositional semantics (Appendix \ref{sec:compositionality_localisation}). Such questions are especially relevant in solving semantic gaps between input prompts and expected responses, as well as localising linguistic knowledge and improving interpretability. One way in which they can be addressed is by systematically assessing the impact of said perturbations on model inference performance, at each model layer. We elaborate on the methodology to achieve this goal in the following section.

\section{Assessing compositional aggregation robustness}
To accurately assess the effects of compositional grouping at different layers of abstraction within transformer models, the inference objective should be a task that is both: 1) strictly dependent on the input tokens and their composition, with few possible input variations; 2) contains as few tokens as possible in the output. For this reason, the following tasks were selected (Figure \ref{fig:cap_method}):\\
\noindent \textbf{1.} Inverse definition modelling (IDM): predicting a term given its definition.\\
\noindent \textbf{2.} Synonym prediction (SP): producing a synonym for a given word.\\
\noindent \textbf{3.} Hypernym prediction (HP): generating a more general term for a given word.\\
\noindent Formal task definitions and input formats are detailed in Appendix \ref{sec:formal_downstream_tasks}.

\noindent \textbf{Constituent-Aware Pooling (CAP).} To introduce compositional perturbations, we propose CAP, a method for pooling (i.e., grouping) LLM activations corresponding to individual tokens into cohesive linguistic units. CAP operates at two levels: (i) word-level: grouping tokens that form a single word, and (ii) phrase-level: grouping tokens that form a single phrase. At the word-level, CAP reverse-maps each model's tokeniser to reconstruct complete words and identify their activation ranges. At the phrase-level, CAP uses a syntactic parser, such as Benepar \cite{kitaev-etal-2019-multilingual, kitaev-klein-2018-constituency}, to align tokens with their corresponding phrasal constituents and define their activation ranges. Further details on the parser evaluation methodology are provided in Appendix \ref{sec:benepar_eval}. \\
\textbf{CAP Pooling Protocols.} CAP is applied progressively across layers using three protocols $\alpha$: \textit{Max:} selects the maximum activation within a segment, identifying dominant features and their propagation through layers; \textit{Mean:} computes the average activation, providing a balanced representation of all token contributions and their collective impact on model decisions; and \textit{Sum:} sums the activations, capturing cumulative information flow and aggregates effects of token interactions. These protocols offer complementary insights into how models process and integrate information: Max reveals feature prominence patterns, Mean shows distributed representation effects, and Sum reflects accumulated semantic content across segments.
\textbf{Transformer conceptualisation and the formalisation of CAP.} This work builds on the mathematical framework of transformers introduced by \cite{elhage2021mathematical}, where computation is formalised into sequential residual blocks. Each layer reads inputs from the residual stream, processes them through its components (attention heads and feed-forward neural networks (FF)), and writes the outputs back into the residual stream. Attention heads are responsible for transferring information between tokens through the self-attention mechanism, allowing each token to attend to others in the sequence. FF apply non-linear transformations independently to each token representation, enhancing the model's expressive capacity. The residual stream stores and propagates information across layers, enabling the integration of new outputs with existing representations while preserving original input information through residual connections. 
Let the transformer model have $L$ layers, input sequence of length $K$, batch size $B$, and inner activations $X$, with with tensor shapes varying by model component as follows:
\begin{itemize}
    \item Attention layers output: $X \in R^{B\times K \times H_m}$, where $H_m$ is the hidden dimension after projection.
    \item FF: $X \in R^{B\times K \times H_f}$, where $H_f$ is the feed-forward dimension.
    \item Residual stream: $X \in R^{B\times K \times H_h}$, where $H_h$ is the hidden dimension. 
\end{itemize}
Let $\mathcal{S}=\{(s_1,e_1), \dots (s_n, e_n)\}$ be the set of syntactic unit ranges (e.g., tokens, words or phrases), where $s_i$ and $e_i$ denote the start and end indices of the $i$-th range. CAP pools/groups activations within these ranges, reducing the sequence dimension $K$ to a grouped dimension $G$, where 
\begin{equation}
    G=K-\Sigma_{i=1}^n(e_i-s_i)
\end{equation}
For each syntactic unit, CAP applies the grouping function $\alpha$ over the range $[s_i, e_i]$ in one of three ways, formalised as follows:
\begin{equation}
    \textbf{Sum:} \quad \alpha([s_i,e_i])= \sum_{t=s_i}^{e_i} X[t]
\end{equation}
\begin{equation}
    \textbf{Mean:} \quad \alpha([s_i,e_i])= \frac{1}{e_i - s_i + 1} \sum_{t=s_i}^{e_i} X[t]
\end{equation}
\begin{equation}
    \textbf{Max:} \quad \alpha([s_i,e_i])= \max_{t \in [s_i,e_i]} X[t]
\end{equation}

The grouped activations transform as follows:
\begin{itemize}
    \item For attention layers output, $X \in R^{B\times K \times H_m}$ becomes $X \in R^{B\times G \times H_m}$.
    \item For FF, $X \in R^{B\times K \times H_f}$ becomes $X \in R^{B\times G \times H_f}$.
    \item For residual stream:, $X \in R^{B\times K \times H_h}$ becomes $X \in R^{B\times G \times H_h}$.
\end{itemize}
This process consolidates activations for each syntactic unit, enabling systematic evaluation of compositional robustness across layers. For simplicity, we demonstrate the operation over these components, but this approach can be extended to any transformer's components, provided that the dimensional requirements for information flow, as described in \cite{elhage2021mathematical}, are respected. For example, consider attention layer internal activations of shape $X \in \mathbb{R}^{B \times H_a \times K \times K}$, where $H_a$ is the number of attention heads, and $K$ represents the query and key token dimensions. Applying CAP with the \textbf{Sum} protocol involves aggregating activations over the query range $[s_i, e_i]$ and the key range $[s_j, e_j]$. The grouped activations are computed as: $\alpha([s_i, e_i], [s_j, e_j]) = \sum_{t=s_i}^{e_i} \sum_{t'=s_j}^{e_j} X[b, h, t, t']$. After applying CAP, the grouped activations have the shape $X \in \mathbb{R}^{B \times H_a \times G \times G}$, where $G$ is the number of grouped syntactic units. This ensures that query-key interactions are consolidated into cohesive syntactic units, aligning activations with higher-level linguistic structures. We examine CAP's reduction ratio (\(K \to G\)) at the word-level and its effects across models, with detailed analysis in Appendix~\ref{sec:token_reduction_analysis}. We refer the reader to Appendix~\ref{sec:cap_position_reduction} for further details on how CAP affects sequence length and interacts with positional encodings.

The CAP effect on models is evaluated by measuring their accuracy post-CAP on a baseline test consisting of examples correctly predicted by the original models. This ensures that the evaluation focuses on instances where CAP directly tests compositional robustness. Specifically, we report three key metrics: the original accuracy ($A_o$), which represents the model's accuracy on the baseline test before applying CAP and establishes a reference for evaluating the grouping effect; the grouped accuracy ($A_c$), which measures the model's accuracy post-CAP, averaged across all CAP protocols (sum, mean, max) and reflects how well the model retains its predictions after compositional grouping; and the accuracy drop ($\Delta A$), defined as $\Delta A = A_o - A_c$, which quantifies the performance loss due to CAP, where lower $\Delta A$ values indicate more robust compositional behaviour and better preservation of semantic information across layers. These metrics offer a framework for comparing tasks and models, allowing a granular assessment of compositional representations.

\section{Empirical analysis}
\subsection{Experimental setup \& datasets}
\paragraph{Datasets and metrics.} The CAP effect is evaluated using three WordNet-derived datasets—definitions, hypernyms, and synonyms—corresponding to the IDM, HP, and SP tasks \cite{fellbaum1998wordnet}. Test examples correctly predicted by the original models ($A_o$) form the baseline for subsequent CAP evaluation. Grouped accuracy ($A_c$) is calculated post-CAP for this subset, ensuring that CAP's effect is isolated to examples where the original models performed correctly. The drop in accuracy ($\Delta A$) is reported per protocol (sum, mean, max) to assess the impact of different aggregation methods on model performance. See Appendix \ref{sec:datasets} for dataset details and Appendix \ref{sec:full_results} for comprehensive results.
% Full dataset details are provided in Appendix \ref{sec:datasets}, and comprehensive accuracy drop results across tasks, protocols, and layers are available in Appendix \ref{sec:full_results}.

% \paragraph{Metrics.}
% To evaluate model performance, we report original accuracy ($A_o$) and accuracy drop after grouping ($\Delta A = A_o - A_c$, where $A_c$ is post-CAP accuracy). These metrics effectively capture the models' performance and the impact of CAP. $\Delta A$ directly quantifies the grouping effect, enabling clear cross-model and cross-task comparisons. An expanded reporting on the results is in Appendix \ref{sec:full_results}.

\begin{table*}[h!]
\centering
\small
\begin{tabular}{|c|c|ccc|ccc|}
\hline
\multirow{2}{*}{\textbf{Model}} & \multirow{2}{*}{\textbf{\begin{tabular}[c]{@{}c@{}}Layer\\Position\end{tabular}}} & \multicolumn{3}{c|}{\textbf{Original}} & \multicolumn{3}{c|}{\textbf{Fine-tuned}} \\
\cline{3-8}
 &  & \textbf{Max} & \textbf{Mean} & \textbf{Sum} & \textbf{Max} & \textbf{Mean} & \textbf{Sum} \\
\hline
\multirow{4}{*}{\textbf{GPT2-large}} 
& 1\%  & 8.06\%   & \cellcolor{red!15}9.15\%  & \cellcolor{green!50!black!10}6.70\% & \cellcolor{red!15}10.61\% & 10.01\% & \cellcolor{green!50!black!10}7.83\% \\
& 25\% & 5.19\%  & \cellcolor{green!50!black!10}4.94\% &  \cellcolor{red!15}5.63\%  & 6.25\% & \cellcolor{green!50!black!10}5.77\% & \cellcolor{red!15}6.32\% \\
& 75\% & \cellcolor{red!15}5.28\%  & 2.62\% &  \cellcolor{green!50!black!10}2.39\%   & \cellcolor{red!15}3.66\% & 1.62\% & \cellcolor{green!50!black!10}0.88\% \\
& 100\%& \cellcolor{red!15}0.84\%  & \cellcolor{green!50!black!10}0.12\%  & 0.19\%   & \cellcolor{red!15}0.22\% & \cellcolor{green!50!black!10}0.16\% & \cellcolor{green!50!black!10}0.16\% \\
\hline
\multirow{4}{*}{\textbf{Gemma-2B}}  
& 1\%  & \cellcolor{red!15}97.91\%   & \cellcolor{green!50!black!10}23.51\%  & 23.75\%  & \cellcolor{red!15}57.58\% & 22.70\% & \cellcolor{green!50!black!10}21.99\% \\
& 25\% & \cellcolor{red!15}86.32\%   & \cellcolor{green!50!black!10}16.20\% & 19.27\%  & \cellcolor{red!15}50.45\%  & \cellcolor{green!50!black!10}14.08\% & 15.57\% \\
& 75\% & \cellcolor{red!15}52.38\%   & 31.03\% & \cellcolor{green!50!black!10}24.74\% & \cellcolor{red!15}21.77\% & 14.99\% & \cellcolor{green!50!black!10}12.80\% \\
& 100\%& \cellcolor{green!50!black!10}6.87\%   & \cellcolor{red!15}10.61\% & \cellcolor{red!15}10.61\% & \cellcolor{red!15}2.21\% & \cellcolor{green!50!black!10}2.05\% & \cellcolor{green!50!black!10}2.05\% \\
\hline
\multirow{4}{*}{\textbf{Qwen-3B}} 
& 1\%& \cellcolor{red!15}12.63\%  & 12.27\%  & \cellcolor{green!50!black!10}11.44\%   & \cellcolor{red!15}7.85\% & 6.71\% &\cellcolor{green!50!black!10} 6.48\% \\
& 25\%& \cellcolor{red!15}18.61\%  & \cellcolor{green!50!black!10}8.59\%  & 9.11\%   & \cellcolor{red!15}10.66\% & \cellcolor{green!50!black!10}4.75\% & 5.82\% \\
& 75\%& \cellcolor{red!15}7.23\%  & 4.00\%  &\cellcolor{green!50!black!10} 3.79\%   & \cellcolor{red!15}3.65\% & 2.83\% & \cellcolor{green!50!black!10}1.85\% \\
& 100\%&\cellcolor{green!50!black!10}0.39\%  & \cellcolor{red!15}0.4\%  & \cellcolor{red!15}0.4\%   & \cellcolor{red!15}0.31\% & \cellcolor{green!50!black!10}0.17\% & 0.2\% \\
\hline
\multirow{4}{*}{\textbf{Llama3-8B}}  
& 1\%  & \cellcolor{red!15}25.49\%   & 24.99\%  & \cellcolor{green!50!black!10}24.94\%  & \cellcolor{red!15}24.44\% & \cellcolor{green!50!black!10}23.42\% & 23.48\% \\

& 25\% & \cellcolor{red!15}20.02\%   & 5.87\% & \cellcolor{green!50!black!10}5.74\%  & \cellcolor{red!15}8.81\%  & 6.03\% & \cellcolor{green!50!black!10}5.92\% \\

& 75\% & \cellcolor{red!15}7.31\%   & \cellcolor{green!50!black!10}3.40\% & 3.54\% & \cellcolor{red!15}5.16\% & 3.47\% & \cellcolor{green!50!black!10}3.29\% \\

& 100\%& \cellcolor{red!15}2.80\%   & \cellcolor{green!50!black!10}1.77\% & \cellcolor{green!50!black!10}1.77\% & \cellcolor{red!15}1.55\% & \cellcolor{green!50!black!10}1.33\% & \cellcolor{green!50!black!10}1.33\% \\
\hline
\end{tabular}
\caption{IDM accuracy drop $\Delta$ in the word-level CAP, highlighting {\color{green!50!black!50}best} and {\color{red!50!black!50}worst} values in both original and fine-tuned models. The layer numbers were normalised to layer positions as percentages of the total layers, which allows comparing equivalent relative depths across models, such as 25\% or 75\% of the total layers, rather than using absolute layer numbers. This method ensures fair comparisons between models, even with different architectures.}
\label{tab:acc_drop_org_finetuned}
\end{table*}

% \paragraph{Datasets}
% Experiments were conducted on three WordNet-derived datasets: definitions, hypernyms, and synonyms \cite{fellbaum1998wordnet}. These correspond to the three tasks: IDM, HP, and SP. Initially, models were tested on these constructed datasets using the correctly predicted samples as the baseline for further analysis (see Appendix~\ref{sec:datasets} for datasets details).
\paragraph{LLMs and evaluated dimensions.}
The methodology was tested across various decoder-only transformer models \cite{vaswani2017attention}. Our main focus was on GPT-2 (small: 124M, medium: 355M, large: 774M parameters) \cite{radford2019language}, Gemma1 (2B parameters) \cite{team2024gemma}, Llama (3B, and 8B parameters) \cite{dubey2024llama}, and Qwen (0.5B, 1.5B, and 3B parameters) \cite{yang2024qwen2}. These models use different tokenisation approaches: byte-level BPE (GPT-2, Qwen), expanded BPE with 128K vocabulary (Llama3), and SentencePiece (Gemma). Models were tested before and after task-specific fine-tuning (3 epochs, learning rate 5e-5). This selection spans diverse architectures, sizes, and tokenisation strategies (see Appendix~\ref{sec:model_params} for further details on the models and fine-tuning parameters).

\paragraph{Experimental setup.}
All experiments were conducted using 2x NVIDIA RTX A6000 and 2x NVIDIA RTX A100 GPUs, with the experimental framework being developed in Python 3.11.5. We used the Transformers (v4.44.2) and PyTorch (v2.4.1) libraries, along with Transformer-lens (v2.6.0), to train and evaluate models and for probing. Benepar (v0.2.0) was used for sentence parsing, and statistical analysis was supported by Scikit-learn (v1.5.2). 

\subsection{Results and discussion}  
\begin{figure*}
    \centering
    \includegraphics[width=.8\linewidth]{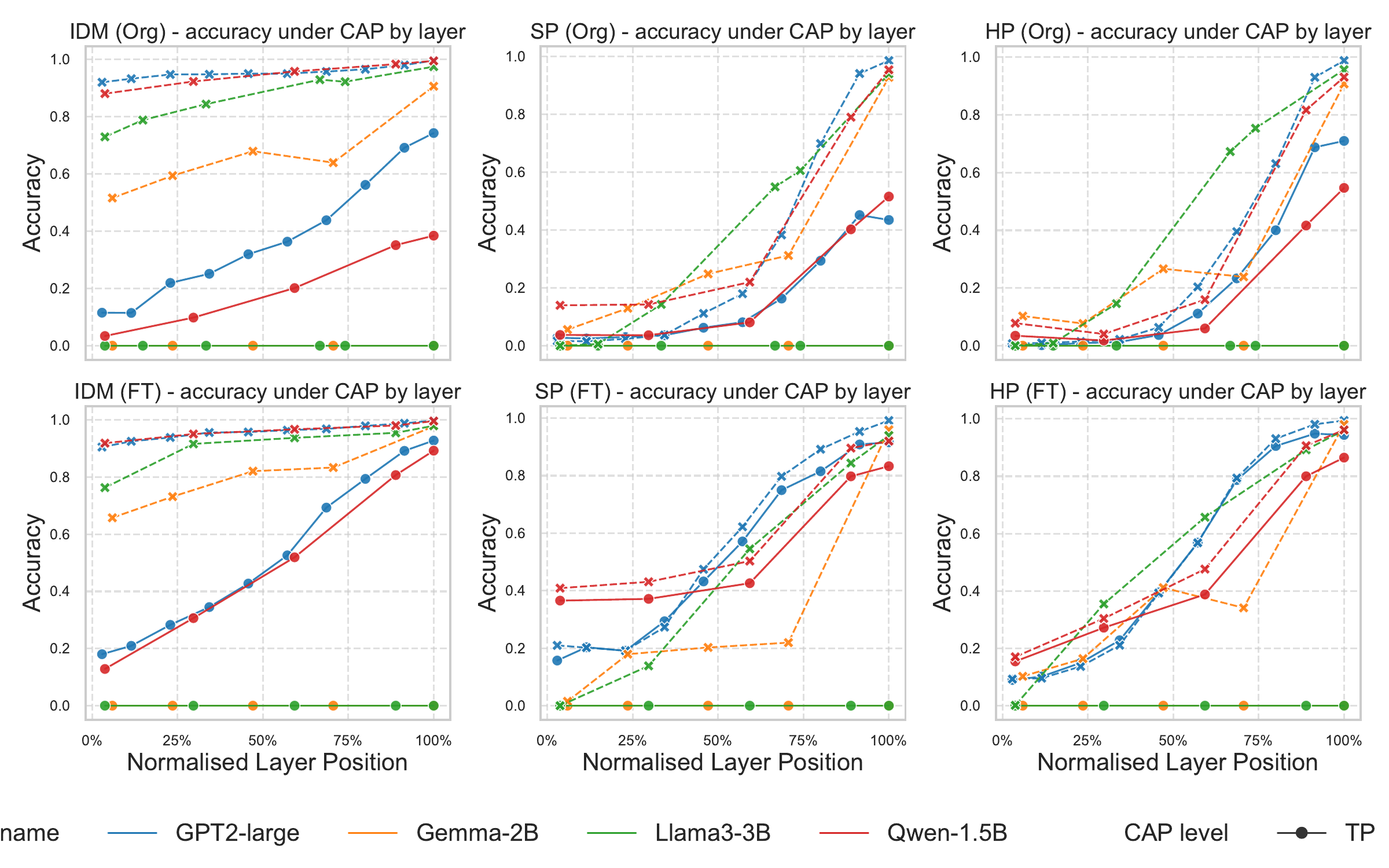}
    \caption{Average grouped accuracy of CAP across different aggregation functions for normalised layer positions (0\%-100\%) is shown for word-level CAP (TW) and phrasal-level CAP (TP). Sub-figures (a)-(c) illustrate the CAP effect on the original (Org) models, while sub-figures (d)-(f) show its impact on the fine-tuned (FT) models. Fine-tuning consistently improves performance, particularly in the middle to late layers (25\%-100\%), while early layers (0\%-25\%) show more variability and lower accuracy across models.}
    \label{fig:avg_group_acc}
\end{figure*}

\textbf{Compositional inference in LLMs is not a purely incremental process.} Contrary to expectations of a smooth and steady layer-wise performance improvement, we observe significant fluctuations when CAP is applied across layers. Performance drops notably in early and middle layers, followed by sharp improvements (Figure~\ref{fig:avg_group_acc} (a)-(c), (e), and (f)), suggesting these layers struggle to process CAPed activations, particularly the pooled linguistic features captured in earlier layers. \textit{Rather than progressively building semantic information from individual tokens to complex phrases, the models appear to focus heavily on \textit{isolated token features}.} 

An important distinction arises between \textbf{TW-CAP}, which groups tokens according to model-specific tokenisation, and \textbf{TP-CAP}, which applies externally parsed syntactic structures. While TP-CAP introduces richer constituent information, it may not align with the model's internal segmentation or syntactic reasoning. This misalignment is not a flaw in CAP, but rather a diagnostic signal: if LLMs encoded human-like syntax, TP-based grouping should be minimally disruptive. The observed drop in performance under TP-CAP suggests that LLMs do not consistently internalise hierarchical syntactic structures. This finding underscores the model's emphasis on local token-level information and supports the conclusions drawn in our information-theoretic analysis.

The results indicate that attention is distributed over input tokens and model layers in a non-systematic and decentralised manner that is highly context-dependent, showing minimal reliance on sequential or positional relationships of constituents. This phenomenon is particularly evident in the sharp decline in SP and HP tasks, where contextual information is limited during phrase-level CAP application. We argue that this behaviour stems from the model's training objective, which maximises information gain in each layer towards predicted tokens at the cost of reducing mutual information between tokens in a single layer. This behaviour means that \textit{aggregation, including syntactic, is performed across multiple layers and thus is not localisable from any single given layer}. An information theoretical analysis elaborates this reasoning in Section~\ref{sec:info_gain_mi}. Our findings highlight how compositional structures are highly sensitive to token representation dynamics across layers, suggesting that performance fluctuations \textit{can be attributed to information loss incurred as a function of token mutual information across layers.}\\
\noindent\textbf{Larger models are more fragile to compositional perturbations.} The IDM task highlights this fragility in larger models, as larger models rely on finer feature extraction. Within families, distinct patterns emerge: original Qwen's smaller variants show better IDM robustness (e.g., at position 25\% there was a 7.69\% drop on Qwen-1.5B vs 12.11\% on Qwen-3B), while Llama3 exhibits capacity-dependent behaviour with the 3B variant being more vulnerable than 8B. Despite having similar reduction ratios to Llama models (see Appendix~\ref{sec:token_reduction_analysis}), Gemma-2B shows greater sensitivity to perturbations (e.g., at position 1\% Max: Gemma-2B drops 97.91\% vs. Llama3-8B's 25.49\%), likely due to its larger vocabulary enabling finer-grained tokenisation. While fine-grained token knowledge benefits standard tasks, it appears to increase susceptibility to compositional perturbations. The superior performance of Llama3-8B over its 3B variant can be attributed to its enhanced capacity for maintaining feature relationships across layers while preserving key compositional information. While larger models excel in standard tasks (see Appendix~\ref{sec:baseline_perf}), \textit{they exhibit a greater reliance on the identification of intrinsic features in the early layers.} We find that phrasal-level CAP substantially impacts Gemma-2B and Llama models, suggesting a heavy dependence on layer-wise information gain, where they separate features in an uncorrelated and highly distinct manner. While this aids in identifying complex feature patterns, it also makes them more vulnerable to contextual noise—\textit{a weakness that threatens their robustness and integrity}. Notably, Qwen models outperform Llama and Gemma despite similar parameter counts, likely due to byte-level BPE tokenisation and multilingual training, which enhance compositional stability, whereas Llama's expanded BPE and Gemma's SentencePiece prioritise efficiency over phrase retention, increasing vulnerability to CAP interventions.\\
\noindent \textbf{Activation abstraction vs the information loss.} Table~\ref{tab:acc_drop_org_finetuned} reveals significant variations in aggregation function performance across sample models for the IDM task (see Appendix~\ref{sec:full_results} for the rest of the models and tasks results). The Max aggregation shows the most dramatic impact. \textit{This finding supports our argument that these models tend to distribute information in a fragmented manner, lacking the integration of compositional (lexical and semantic) information across tokens and contiguous layers.} The Mean aggregation provides more balanced results, though performance drops still indicate \textit{absence of consistent compositional mechanisms.} This issue becomes more pronounced in token-phrases experiments (Figure~\ref{fig:avg_group_acc}). The \textit{Sum aggregation consistently outperformed other methods}, with Mean aggregation following closely behind, particularly in original models. The Sum aggregation reflects the cumulative effect of aggregating tokens into larger segments, reinforcing our earlier conclusion. Instead of progressively building semantic information across layers, \textit{the models exhibit cumulative information loss,} particularly when interventions occur in early layers.\\
\noindent\textbf{Fine-tuning enhances recovery capabilities across models.} Figure \ref{fig:avg_group_acc} (d-f) demonstrates improved performance maintenance post-fine-tuning across all model families, with strongest gains in 75\%-100\% layer positions. SP tasks showed maximum benefit, attributed to high task specificity and minimal activation reduction under CAP. Max aggregation displayed the greatest improvement post-fine-tuning, likely due to enhanced retention of key information. For instance, Gemma-2B's accuracy drop decreased from 97.91\% to 57.65\% in the 1\% layer, while Qwen-3B improved from 7.23\% to 3.65\% in the 75\% layer. Mean aggregation benefits were also substantial in smaller models, with Gemma-2B's 75\% layer drop reducing from 31.03\% to 15.00\%. The Qwen family showed consistent improvements across all aggregation types, though smaller models like GPT2-large demonstrated minimal gains, suggesting potential overfitting. Notably, larger models like Llama3-8B showed minimal gains from fine-tuning in IDM tasks, indicating that standard fine-tuning objectives may not directly enhance compositional robustness. Although fine-tuning strengthens models' resilience under CAP, it does not fully resolve the challenge of forming stable compositional semantic representations, highlighting an architectural limitation in current transformer models.

% \noindent\textbf{Fine-tuning expedites the recovery process.}
% As seen in Figure \ref{fig:avg_group_acc} (d-f) and Table \ref{tab:acc_drop_org_finetuned}, there is a positive correlation between the models' ability to maintain performance and fine-tuning (FT). The biggest improvements are observed in the 75\%-100\% layer positions. SP demonstrated the best gain from FT, which we attribute to the task's high specificity and the minimal activation reduction under CAP. Max aggregation shows the greatest improvement with FT, likely due to fine-tuning amplifying dominant activations and helping retain key information. For example, in Gemma-2B, the drop decreased from 97.91\% to 57.65\% in the 1\% layer, and in Llama3-8B, from 14.98\% to around 7\% in the 75\% layer. Mean aggregation also benefited from FT, with the accuracy drop in Gemma-2 B's 75\% layer reduced from 31.03\% to 15.00\%. However, except for Gemma-2B, Sum aggregation saw only slight improvements. For smaller models like GPT2-large, results showed either the opposite behaviour or minimal improvement, which we believe is due to overfitting. While fine-tuning speeds up recovery, it does not fully mitigate performance reduction, indicating that current Transformer architectures still struggle to form compositional semantic representations from individual tokens.

\section{Information Gain \& Token Mutual Information}\label{sec:info_gain_mi}
The empirical findings can be explained by looking at the autoregressive next-token objective of a transformer model from an information theoretical standpoint: examining the relationship between each generated token $Y$ to the input token representations $R_l(X)$ of each layer $l$, in terms of Information Gain $IG_{Y, R_l(X)}$, and the aggregation of a pair of input token representations $R_l(X_i), R_l(X_j)$ in terms of their Mutual Information $I(R_l(X_i), R_l(X_j))$.

$IG_{Y, R_l(X)}$ quantifies the amount of information gained about the predicted token $Y$ from the observation of the $R_l(X)$, for which the expectation is the mutual information $I(Y, R_l(X))$ of $Y$ and $R_l(X)$, which is equivalent to the reduction in entropy of $Y$ achieved by learning the state of $R_l(X)$: $IG_{Y, R_l(X)}(Y, r) = H(Y) - H(Y | r)$.

During training, $R_l(X)$ will be adjusted in a way that reduces the uncertainty about $Y$, meaning it will promote the maximisation of $IG_{Y, R_l(X)}$ for any given layer $l$, which can be expressed as:

\begin{equation}\label{eq:info_gain_pred}
    IG_{Y, X} = max(\sum_{l}{IG_{Y, R_l(X)}})
\end{equation}

\noindent where $IG_{Y, X}$ represents the information gain of $Y$ w.r.t. input token $X$.

When looking at two input tokens $X_i, X_j$, the higher the mutual information $I(R_l(X_i), R_l(X_j))$ is, the lower the impact that aggregating $R_l(X_i)$ and $R_l(X_j)$ would have over $IG_{Y, X}$, as those variables share more of the same information. Intuitively, that would apply to linguistic composition, e.g., tokens that form a word and thus have a stronger dependence when observed together. 

However, as the model's ability to predict $Y$ is contingent on the accumulated information of all layers, and Equation~\ref{eq:info_gain_pred} is independent of layer order, there is an intrinsic incentive to delay the aggregation of information (to later layers), as
\begin{equation}\label{eq:info_gain_repr}
    IG_{R_{l_p}(X), R_{l_q}(X)} < IG_{R_{l_p}(X), R_{l_r}(X)}, \quad \forall p < q < r,
\end{equation}

where $p$, $q$ and $r$ are layer indices, i.e., subsequent layers have more information about the inputs than previous ones. This can be explained in that optimising Equation~\ref{eq:info_gain_pred} can be achieved by retaining at each $R_{l_p}(X)$ only the necessary information to maximise $\sum_{i, j}{IG_{R_{l_q}(X_i), MHA(R_{l_p}(X_j))}}$, where $MHA(R_{l_p}(X_j))$ is the multi-head attention weighted representation. Such an objective implies minimising the mutual information $I(R_{l_p}(X_i), R_{l_p}(X_j))$, i.e., reducing redundancy across tokens from the same layer. Therefore, token dependencies will tend to be modelled by aggregation paths spanning multiple layers, with more layers allowing for more complex and longer paths. This is in line with the findings of Mechanistic Interpretability studies \cite{elhage2021mathematical, conmy2023towards}. Equation~\ref{eq:info_gain_repr} also implies that the earlier an aggregation is done, the larger the impact it will have on $IG_{Y, X}$, which explains the empirical results. The effects of $I(R_l(X_i), R_l(X_j))$ on LLMs are further compounded by the tokenisation objective (e.g., BPE, WordPiece), which \textit{minimises} $I(X_i, X_j)$, i.e., token redundancy, as a means of reducing the vocabulary size, leading to longer aggregation paths.

\section{Related work}
Compositionality, the principle that the meaning of complex expressions is derived from their parts and structure, is foundational in linguistics, cognitive science, and AI \cite{fodor1975language, montague1975formal, tull2024towards}. In neural models, compositionality enables generalisation and interpretability, yet remains difficult to diagnose and enforce \cite{Compositionality_Donatelli}. Several studies investigate how and where compositional representations emerge in transformer models. \citet{carvalho2022montague} observed similar effects in adjective-noun phrase probing, while \citet{haslettmuch2024} found that models struggle to segment or represent morphemes, especially in non-Latin scripts, suggesting breakdowns in both form and meaning composition. The logit lens \cite{nostalgebraist2020logit} demonstrated that transformers build predictions progressively where early layers make initial guesses and deeper layers refine guesses with broader context. \cite{dai-etal-2022-knowledge} show feed-forward layers act as key-value memories, combining information for complex predictions. MEMIT \cite{meng2023memit} and PMET \cite{li2024pmet} show how controlled inferences can be built by manipulating models' components. Some nuance emerges in later-layer behaviours. DecompX~\cite{modarressi2023decompx} traced token representations layer-by-layer and observed partial shifts toward integration. \citet{yu2020assessing} tested model encoding and found that transformers mainly encode individual word content rather than true phrase-level meaning. While some models appear more compositional under certain conditions, general trends remain unclear. For example, \citet{dankers_paradox_2022} demonstrate that models can show unexpectedly high or low compositionality depending on the data and task, suggesting exposure and framing affect outcomes as much as architecture. \citet{petty_impact_2024} show that deeper Transformers tend to generalise more compositionally than shallower ones, though the benefits diminish beyond a certain depth. This highlights that architectural depth, not just scale, may shape compositional ability, though with diminishing returns. In multi-step reasoning tasks, models often fall back on shallow pattern matching rather than true decomposition \cite{dziri2023faith}.
% \citet{dziri2023faith} tested LLMs on multi-step reasoning tasks and found that, rather than truly decomposing problems, these models often match inputs to memorised patterns or linearised subgraph fragments.

Prior work has primarily relied on synthetic tasks to assess compositional generalisation, focusing on properties such as systematicity, productivity, and substitutivity \cite{ijcai2020p708, pmlr-v80-lake18a}, these setups often abstract away from the complexities of natural language. More recent studies using natural data are often limited to small domains such as semantic parsing or machine translation \cite{pmlr-v80-lake18a, kim-linzen-2020-cogs}, and typically lack insight into internal representations.

In contrast to prior works focused on final outputs or synthetic tasks, CAP is a method for probing compositional structure within LLMs using real inputs. It intervenes directly on hidden activations, merging token-level representations into word- or phrase-level constituents at various depths. This allows us to evaluate where semantic composition occurs and how robust LLMs to structured perturbations. Unlike surface-level probes, CAP provides a targeted, activation-level lens on how meaning is constructed and distributed across model layers and linguistic units.

\section{Conclusion}
This work systematically analyses the robustness of transformer-based LLMs to compositional perturbations. Motivated by studies highlighting an unexpected gap between linguistic compositionality and LLM representations, we characterised the impact of compositional aggregation at each inference step and provided an information-theoretical explanation. Our findings indicate a pattern where token dependencies are modelled by aggregation paths spanning multiple layers, and complex token structure learning comes at the cost of higher sensitivity to perturbations at inputs and earlier layers. Based on the relation between information gain from input to predicted token and mutual information between token representations, we postulate that compositional semantic representations cannot be isolated to any particular (intermediate) stage of a standard transformer model. These insights suggest that future compositional-aware models should explore specialised architectures or training objectives. Natural extensions include analysing encoder-based and encoder-decoder transformers and investigating final token representations to further understand internal compositional mechanisms.

\section*{Limitations}
Several limitations are acknowledged in our paper. First, the WordNet dataset may not fully represent language diversity across all domains. Second, the employed transformer models are decoder-based only and could be subject to biases from their training data. Third, our findings depend on the Benepar parsing model, which may introduce inaccuracies in linguistic analysis. Additionally, while our tasks provide an indirect signal of meaning preservation, incorporating explicit reconstruction tasks in future work could offer complementary insight into how CAP affects the retention of input-level information. Finally, the applicability of our results to other languages has not been tested. Expanding CAP to multilingual settings and testing with alternative parsers or models trained with different positional encodings would further validate the generality of our findings.

\section*{Ethical Statement}

The proposed framework aims to have a positive impact on improving the critical understanding of the mechanisms involved in language interpretation in transformers. A more complete understanding of these mechanisms requires coordination with other interpretability methods.

\bibliography{references}

\appendix
\section{Compositionality and Localisation}\label{sec:compositionality_localisation}
The concept of linguistic compositionality has evolved from its origins in Frege's work \cite{Frege1892}, which started conceptualising the notion that the meaning of a complex expression is determined by its constituent parts and their syntactic arrangement. This principle was formalised by Montague \cite{Montague1970a, Montague1970b}, who applied mathematical rigour to natural language semantics, thereby reinforcing the compositional approach within formal semantics. Linguistic phenomena such as idioms, context-dependence, and metaphor, which seemed to violate compositionality, prompted debates on its universality \cite{KatzPostal1963, Jackendoff1997}, with theoretical accounts evolving to integrate these phenomena, leading to a more nuanced understanding that balances strict compositional rules with allowances for non-compositional elements \cite{Partee1984}. 

While the syntactic-logical connection entailed by formal models is not assumed to be induced by neural language models, there is a common assumption that those models should entail a syntactic compositionality function, which allows for a systematic model for meaning composition, i.e., that the syntactic structure of a complex expression \( s \) is significantly determined by the syntactic properties of its constituent parts and the rules used to combine them. Formally, for any sentence \( s \), its syntactic properties can be defined as a function \( f \) of the syntactic properties of its immediate constituents \( s_1, s_2, \dots, s_n \) and the syntactic operations applied:

\begin{equation}
\begin{split}
    \text{Syntax}(s) = f\left(\text{Syntax}(s_1), \text{Syntax}(s_2), \dots, \right. \\
    \left. \text{Syntax}(s_n), \text{Rules}\right)
\end{split}
\end{equation}

Within the context of distributed representations, a meaning representation can be factored into its syntactic and content (term embedding) components. A compositional distributional semantic model merges syntactic compositionality with distributional semantics by representing token meanings as vectors (token embeddings) in a continuous semantic space and combining them according to syntactic structure. Formally, each token \( t \) is associated with a vector \( \mathbf{v}_t \in \mathbb{R}^n \) that captures its semantic content based on distributional information.

For a complex syntactic expression \( s \) composed of constituents \( s_1, s_2, \dots, s_n \), the semantic representation \( \mathbf{v}_s \) is computed using a compositional function \( f \) that integrates both the vectors of the constituents and the syntactic operations applied:

\begin{equation}
\mathbf{v}_s = f\left( \mathbf{v}_{s_1}, \mathbf{v}_{s_2}, \dots, \mathbf{v}_{s_n}, \text{Syntactic structure} \right)
\end{equation}

This function \( f \) is designed to reflect syntactic compositionality by structurally combining the embeddings of the constituents according to the syntactic rules governing their combination. 

In the context of a specific transformer-based LM model implementing an interpretation function of an input s, the question which is central to this work is whether the contiguous composition of tokens is reflected within the structure of the transformer-based LMs and its constituent parts, layers  \(l_0 ... l_n\), multi-head attention, feedforward layers and residual connections, i.e. whether the representations \( \mathbf{h}_i^{(k)} \) at each layer \( l_k \) explicitly encode the composition of contiguous tokens \( t_i, t_{i+1} \), and how the model's components contribute to this encoding.
\section{Elaborations on Experimental Setup}
\subsection{Downstream Task Definitions}\label{sec:formal_downstream_tasks}
The tasks selected for this study are designed to evaluate the effects of compositional aggregation, focusing on tasks that are strictly dependent on input tokens and their compositional semantics while minimising variability. Each task produces a single-token output, and predictions are considered correct if they exactly match the target token. The following are the formal definitions for each task.

\textbf{Inverse Definition Modelling (IDM):}  
The \textit{IDM} task involves predicting a term \( T \) based on a given natural language definition \( D \). Let \( D = \{d_1, d_2, \dots, d_n\} \) represent the sequence of tokens constituting the definition. The goal is to generate the corresponding term \( T \), where:
\begin{equation}
T = \arg\max_{t \in \mathcal{V}} P(t \mid D)
\end{equation}
Here, \( \mathcal{V} \) is the vocabulary of possible terms, and \( t \) is a candidate term. A prediction is correct if the term \( T \) exactly matches the target term.
The task prompt used for IDM was structured as follows:
\begin{center} \texttt{"<definition> is called a"} \end{center} For example, given the definition "A domesticated carnivorous mammal that typically has a long snout, an acute sense of smell, non-retractile claws, and a barking or howling voice," the task would require the model to predict the term "dog."

\textbf{Synonym Prediction (SP):}  
The \textit{SP} task requires the model to generate a synonym \( S \) for a given word \( W \). Let \( W \in \mathcal{V} \) represent the input word. The task is to predict a synonym \( S \), such that:
\begin{equation}
S = \arg\max_{s \in \mathcal{V}} P(s \mid W)
\end{equation}
where \( s \) is a candidate synonym from the vocabulary \( \mathcal{V} \). The prediction is considered correct if \( S \) exactly matches the target synonym.
The task prompt used for SP was structured as follows:
\begin{center} \texttt{"<word> is a synonym of"} \end{center} For instance, given the input word "happy," the task would ask the model to predict the synonym "joyful."

\textbf{Hypernym Prediction (HP):}  
The \textit{HP} task involves predicting a more general term, or hypernym, \( H \) for a given word \( W \). Let \( W \in \mathcal{V} \) represent the input word. The objective is to predict a hypernym \( H \), such that:
\begin{equation}
H = \arg\max_{h \in \mathcal{V}} P(h \mid W)
\end{equation}
where \( h \) is a candidate hypernym. The prediction is correct if \( H \) exactly matches the intended hypernym.
The task prompt used for HP was structured as follows:
\begin{center} \texttt{"<word> is a type of"} \end{center} For example, given the word "cat," the task would ask the model to predict the hypernym "animal."

These tasks focus on generating precise, single-token predictions, allowing for a rigorous evaluation of the model's ability to capture and process compositional semantics.

\subsection{Dataset Descriptions and Preprocessing}
\label{sec:datasets}
The training and test datasets are constructed by extracting definitions, hypernyms, and synonyms for each synset from WordNet \cite{fellbaum1998wordnet}, whose usage is unencumbered by licensing restrictions. WordNet is a lexical database of the English language, containing over 117,000 synsets of nouns, verbs, adjectives, and adverbs. Each synset represents a unique concept and is annotated with part of speech, definition, hypernyms, synonyms, and other semantic relationships. It is focused on general-purpose vocabulary and does not target specific demographic groups or domains. Definitions were cleaned using typical preprocessing techniques, such as removing special characters, punctuation, and extra spaces, and removing parenthesised content when necessary. The dataset was initially split 80-20, with 20\% used for training. The remaining 80\% was then split 90-10, with 10\% for validation and 90\% for testing. The test dataset was filtered to retain only single-token predictions matching each model's tokenisation. Table~\ref{tab:test_set_sizes} shows the test dataset sizes used for each task and model, including inverse dictionary modelling (IDM), synonym prediction (SP), and hypernym prediction (HP).

\begin{table}
\centering
\resizebox{\columnwidth}{!}{%
\begin{tabular}{|l|c|c|c|}
\hline
\textbf{Model} & \textbf{Task} & \textbf{Original Test Set} & \textbf{Fine-tuned Test Set} \\
\hline
\multirow{3}{*}{GPT2 (S,M,L)} & IDM & 11,948 & 8,651 \\ 
                              & SP  & 7,753  & 5,578 \\ 
                              & HP  & 25,364 & 18,273 \\
\hline
\multirow{3}{*}{Gemma-2B}     & IDM & 24,831 & 17,859 \\ 
                              & SP  & 16,014 & 11,533 \\ 
                              & HP  & 44,687 & 32,209 \\ 
\hline
\multirow{3}{*}{Llama3 (3B, 8B}    & IDM & 14,991 & 10,828 \\ 
                              & SP  & 9,360  & 6,723 \\ 
                              & HP  & 31,962 & 23,070 \\
\hline
\multirow{3}{*}{Qwen2.5 (0.5B, 1.5B, 3B)}   & IDM & 14,927 &  10,780\\ 
                                            & SP  & 9,195 & 6,598 \\ 
                                            & HP  & 31,845 & 23,000 \\
\hline
\end{tabular}%
}
\caption{Test set sizes for each model and task (IDM: Inverse Dictionary Modelling, SP: Synonym Prediction, HP: Hypernym Prediction) derived from WordNet.}
\label{tab:test_set_sizes}
\end{table}
\begin{table}
\centering
\resizebox{\columnwidth}{!}{%
\begin{tabular}{|l|c|c|c|c|c|c|}
\hline
\textbf{Model} & \textbf{Params} & \textbf{Layers} & \textbf{D\textsubscript{model}} & \textbf{Heads} & \textbf{Act.} & \textbf{MLP Dim} \\
\hline
GPT2-small & 124M & 12 & 768 & 12 & GELU & 3072           \\
GPT2-medium& 302M& 24 & 1024& 16 & GELU & 4096           \\
GPT2-large & 708M& 36 & 1280& 20 & GELU & 5120           \\
Gemma-2B &  2B   & 32 & 4096 & 16 & GELU & 8192           \\
LLama3-3B & 3.2B & 28 & 3072 & 24 & SiLU & 8192          \\
LLama3-8B & 7.8B & 32 & 4096 & 32 & SiLU & 14336          \\
% LLama3-8B (Instruct) & 7.8B & 32 & 4096 & 32 & SiLU & 14336  \\
Qwen2.5-0.5B & 391M & 24 & 896 & 14 & SiLU & 4864          \\
Qwen2.5-1.5B & 1.4B & 28 & 1536 & 12 & SiLU & 8960          \\
Qwen2.5-3B   & 3.0B & 36 & 2048 & 16 & SiLU & 11008          \\
\hline
\end{tabular}%
}
\caption{Model properties across architectures. Params: number of parameters, Layers: number of layers, D\textsubscript{model}: size of word embeddings and hidden states, Heads: number of attention heads, Act.: Activation function, MLP Dim: dimensionality of the FF layers.}
\label{tab:model_properties}
\end{table}

\begin{table}[h]
\centering
\small
\resizebox{\columnwidth}{!}{%
\begin{tabular}{|c|ccc|ccc|}
\hline
\multirow{2}{*}{\textbf{Model}} & \multicolumn{3}{c|}{\textbf{Original}} & \multicolumn{3}{c|}{\textbf{Fine-tuned}} \\
\cline{2-7}
& \textbf{IDM} & \textbf{SP} & \textbf{HP} & \textbf{IDM} & \textbf{SP} & \textbf{HP} \\
\hline
GPT2-small   & 7.10\%  & 2.59\%  & 17.04\% & 13.52\%
  & 8.18\%  & 26.59\% \\
\hline
GPT2-medium  & 10.70\%  & 4.27\%  & 16.77\% & 16.34\%  & 11.65\%  & 28.75\% \\
\hline
GPT2-large   & 11.33\%  & 5.93\%  & 13.90\% & 17.80\%  & 11.78\%  & 27.66\% \\
\hline
Gemma-2B     & 16.76\%  & 6.38\% & 10.16\% & 9.57\% & 10.75\% & 23.31\% \\
\hline
Llama3-8B    & 25.17\%  & 10.80\% & 15.30\% & 18.28\% & 10.75\% & 24.14\% \\
\hline
% Llama3-8B (Instruct)    &1.61\%  & 8.41\% & 12.09\% & 19.36\% & 10.97\% & 24.73\% \\
% \hline
Llama3-3B    & 20.51\%  & 8.26\% & 12.19\% & 26.42\% & 13.43\% & 31.1\% \\
\hline
Qwen-0.5B    & 8.21\%  & 6.10\% & 12.03\% & 18.83\% & 10.94\% & 28.03\% \\
\hline
Qwen-1.5B    & 12.35\%  & 7.61\% & 14.64\% & 30.01\% & 13.70\% & 31.31\% \\
\hline
Qwen-3B    & 13.35\%  & 7.53\% & 14.40\% & 31.80\% & 13.66\% & 31.95\% \\
\hline
\end{tabular}%
}
\caption{Baseline performance of various models on three tasks: (inverse dictionary modelling) IDM, synonym prediction (SP), and hypernym prediction (HP). The values represent the accuracy of each model's original and fine-tuned versions. }
% Asterisks indicate .}
\label{tab:baseline_performance}
\end{table}

\subsection{Model Specifications and Fine-tuning Parameters}
\label{sec:model_params}
Table \ref{tab:model_properties} provides a comparative overview of various Transformer models used in this study. We used GPT2 models (released under the Modified MIT License), Gemma-2B (released under the Gemma Terms of Use), Llama3 models (released under the Meta Llama 3 Community License), and Qwen models (released under Apache License 2.0). The used models were mainly pre-trained on English data, with Qwen and LLama models providing additional multilingual support, which is English, German, French, Italian, Portuguese, Hindi, Spanish, and Thai for LLama, and more than 10 languages, including Chinese, English, French, Spanish, Portuguese, Russian, Arabic, Japanese, Korean, Vietnamese, Thai, and Indonesian for Qwen. All models were used for research purposes, specifically for language modelling and text generation in English, aligning with their intended usage. The models differ in their number of parameters, layers, heads, and feedforward (FF) dimensions. The number of parameters ranges from 85M for GPT2-small to 7.8B for LLama3-8B. The activation functions and FF dimensions also highlight variations in the internal processing architecture, influencing the models' performance across different tasks. In addition to these architectural differences, the models were fine-tuned using a consistent set of hyperparameters. The fine-tuning process spanned over three training epochs with a batch size of 16. The learning rate was set to 5e-5, while a weight decay of 0.01 was applied to prevent overfitting. Training logs were generated every 200 steps, with model checkpoints saved every 1000 steps, but limited to retaining only one checkpoint to manage storage efficiently. The evaluation strategy during fine-tuning was set to evaluate at the end of each epoch, and similarly, the model was saved at the end of each epoch as well.

\subsection{Handling of Sequence Reduction and Positional Encoding in CAP}\label{sec:cap_position_reduction}
% One aspect we would have liked the paper to address is how it handles the change in sequence length. By the very nature of the proposed methodology, CAP reduces the number of tokens in the sequence, shortening the input from K to G. If we are not mistaken, the paper does not explicitly discuss how it deals with the reduction in the number of tokens. Does the model simply shorten the sentence and pad the remaining positions with zeros? Maybe keeping the aggregated encoding of group i in the ei position and then apply masking? Another concern is how CAP affects positional embeddings. Since Transformers rely on these for structure, reducing the sequence length may misalign positional information, potentially contributing to the observed performance drop. This is a worrying factor that should be clarified.”
CAP reduces the number of token-level activations from the original input length \( K \) to a shorter grouped sequence length \( G \), by merging activations corresponding to word-level or phrase-level constituents. This reduction is applied post-token embedding and affects intermediate activations within the transformer, specifically the outputs of residual blocks or their internal components (e.g., attention or feedforward sublayers). From the point of CAP application onward, the model processes a reduced-length sequence of size \( G \). This operation does not alter the model’s input embeddings or positional encodings.

\paragraph{Effect of Positional Encoding Schemes.} The impact of this reduction depends on the positional encoding strategy used by the model: (i) \textbf{GPT2 models} use \textit{Sinusoidal positional embeddings}, where each position index corresponds to a unique learned embedding. While CAP does not alter these embeddings directly, reducing the sequence length at intermediate layers can introduce misalignment between positional indices and semantic content. This may disrupt downstream attention or feedforward computations that assume consistent positional context; (ii) \textbf{LLaMA, Qwen, and Gemma} models use \textit{rotary positional encodings (RoPE)}, which encode position relationally through rotation in embedding space. These relative encodings are more robust to changes in sequence length, and CAP has a milder impact on positional semantics in these models. Nevertheless, changes in sequence structure may still affect how models integrate cross-token context.

Although CAP does not interfere with the model's input or positional embedding layer, it alters the spatial structure of activations mid-forward pass. This may influence how transformers aggregate information across positions, especially in models with absolute position encoding. Nevertheless, we did not observe severe performance degradation in those models compared to others. We acknowledge this as a potential contributing factor to the observed degradation under CAP and consider it an important area for future study. 

Namely, Embedding-level analysis represents a promising direction for future exploration. Although this work evaluates a wide range of models with differing positional encoding schemes, we acknowledge the need for more targeted analysis of how CAP interacts with these embeddings. In particular, it would be valuable to quantify the impact of CAP under controlled conditions that isolate embedding effects. For instance, experiments using fixed or masked positional encodings, or applying CAP to models trained from scratch with alternative positional schemes, could help disentangle the influence of compositional pooling from that of positional structure.

\section{Token Reduction Analysis}\label{sec:token_reduction_analysis}
Table~\ref{tab:reduction_percentages} presents an analysis of activation reduction percentages across different LLMs, particularly for the token-to-words case. In this context, the mean represents the average reduction percentages across samples, while the standard deviation indicates the variability of these reductions. While models within a family (e.g., Qwen) share the same tokeniser and vocabulary, the reduction percentages still vary across tasks (e.g., SP vs. HP) because different tasks involve input definitions or prompts with different average sentence lengths and syntactic complexity, which in turn affect how many groupings are formed under CAP. In other words, although the tokeniser is fixed, the number and size of groupable units (e.g., multi-token words or phrases) are input-dependent. The purpose is to assess whether token reduction across models would highly influence the results of CAP. 
\begin{table}
\centering
\small
\begin{tabular}{|l|c|c|}
\hline
\textbf{Model} & \textbf{Task} & \textbf{Mean ± Std} \\
\hline
\multirow{3}{*}{GPT2 (S)} & IDM & $3\pm 5$  \\ 
                          &SP   & $27\pm 9$   \\ 
                          & HP  & $27\pm10$  \\
\hline
\multirow{3}{*}{GPT2 (M)} & IDM & $3\pm 5$  \\ 
                          &SP   & $28\pm10$   \\ 
                          & HP  & $26\pm11$  \\
\hline
\multirow{3}{*}{GPT2 (L)} & IDM & $3\pm 5$  \\ 
                          &SP   & $27\pm9$   \\ 
                          & HP  & $26\pm11$  \\
\hline
\multirow{3}{*}{Gemma-2B}     & IDM & $9\pm4$  \\ 
                              & SP  & $19\pm9$ \\ 
                              & HP  & $30\pm9$ \\ 
\hline
\multirow{3}{*}{Llama3-3B}    & IDM & $10\pm5$ \\ 
                              & SP  & $23\pm6$  \\ 
                              & HP  & $28\pm6$ \\
\hline
\multirow{3}{*}{Llama3-8B}    & IDM & $10\pm5$ \\ 
                              & SP  & $21\pm7$  \\ 
                              & HP  & $28\pm9$ \\
\hline
% \multirow{3}{*}{Llama3-8B instruct}    & IDM & $13\pm4$ \\ 
%                               & SP  & $30\pm6$  \\ 
%                               & HP  & $28\pm8$ \\
% \hline
\multirow{3}{*}{Qwen 0.5B}    & IDM & $3\pm5$ \\ 
                              & SP  & $9\pm11$  \\ 
                              & HP  & $20\pm10$ \\
\hline
\multirow{3}{*}{Qwen 1.5B}    & IDM & $3\pm5$ \\ 
                              & SP  & $12\pm10$  \\ 
                              & HP  & $19\pm10$ \\
\hline
\multirow{3}{*}{Qwen 3B}    & IDM & $3\pm5$ \\ 
                              & SP  & $12\pm10$  \\ 
                              & HP  & $19\pm10$ \\
\hline
\end{tabular}%
\caption{Reduction percentages}\label{tab:reduction_percentages}
\end{table}
\paragraph{Token reduction is a factor but not the sole determinant of performance degradation.} The results presented in Tables~\ref{tab:full_results_tw_IDM}, ~\ref{tab:full_results_tw_SP}, and~\ref{tab:full_results_tw_HP} indicate that while token reduction percentage influences performance degradation, it is not the sole determining factor. Several key observations support this conclusion, which is discussed below. 

First, we observe that higher token reduction does not always lead to a greater performance drop. For instance, models such as Gemma-2B and Llama3-8B exhibit high token reduction percentages (Table~\ref{tab:reduction_percentages}), yet their performance degradation varies significantly across tasks and layer positions. Also, despite lower token reduction percentages, the models Qwen 0.5B and GPT2-small still show substantial accuracy drops, particularly in early layers in the SP and HP tasks. Second, model size and depth influence degradation, as evident in the larger models (e.g., Llama3-8B, Gemma-2B) exhibiting greater fragility to CAP interventions, particularly in early layers (1\% and 25\%). Third, as discussed in the paper, layer-specific variability suggests hierarchical processing differences. Early-layer CAP interventions cause severe accuracy drops in large models but have a less pronounced effect in smaller models, suggesting that deeper architectures defer compositional integration to later layers. Further, fine-tuning reduces degradation in later layers (75\% and 100\%), implying that learned representations in deeper layers mitigate the effects of early perturbations. Finally, architectural differences influence sensitivity. We observe that higher MLP dimensions (e.g., Llama3-8B: 14,336 vs. GPT2-small: 3,072) correlate with greater vulnerability to CAP perturbations, likely due to increased parameter redundancy and disruption of the key-value recall mechanism in MLPs \cite{meng2022locating}.   

While the token reduction percentage contributes to performance degradation, it is insufficient to fully explain the observed variations. Task nature, model size, layer depth, activation functions, and MLP dimensions collectively influence the robustness of CAP interventions. Larger, deeper models demonstrate greater sensitivity to early perturbations, while fine-tuning helps recover performance in later layers. These findings suggest that effective compositional representations in LLMs are distributed rather than localised, requiring specialised architectures or training objectives to improve robustness.

% \begin{table}[h]
%     \centering
%     \caption{Mean Reduction Percentage Across Models for TW Logs}
%     \label{tab:reduction_results}
%     \begin{tabular}{|c|c|c|}
%         \hline
%         \textbf{Model} & \\textbf{Mean Reduction (\%)} & \textbf{Standard Deviation} \\
%         \hline
%         GPT2 Small & X.XX & X.XX \\
%         GPT2 Medium & X.XX & X.XX \\
%         GPT2 Large & X.XX & X.XX \\
%         Gemma 2B & X.XX & X.XX \\
%         Llama 3B & X.XX & X.XX \\
%         Llama 8B & X.XX & X.XX \\
%         Llama 8B instruct & X.XX & X.XX \\
%         \hline
%     \end{tabular}
% \end{table}
\begin{figure*}[]
    \includegraphics[width=0.95\textwidth]{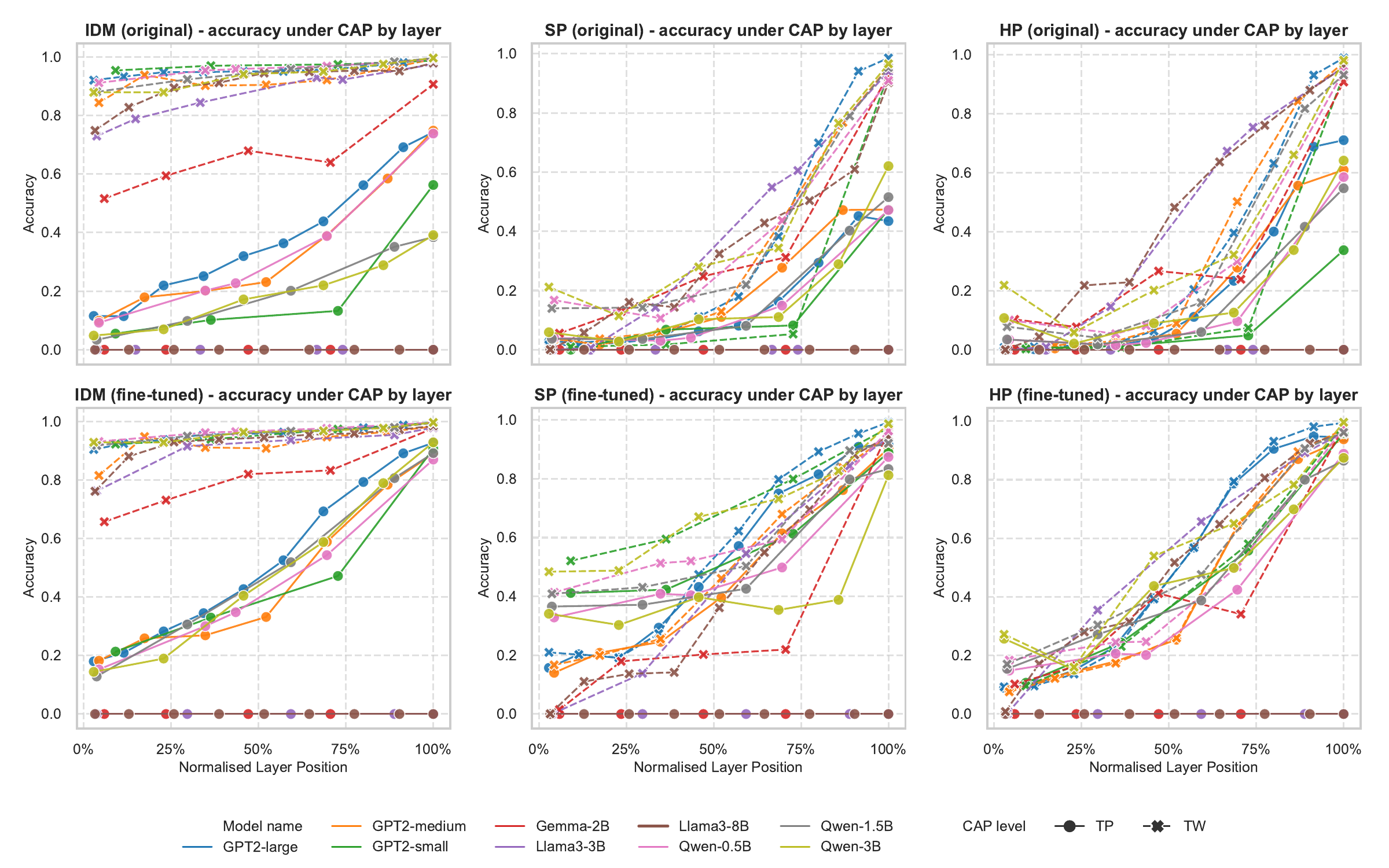}
    \caption{Average grouped accuracy of CAP across different aggregation functions for normalised layer positions (0\%-100\%) is shown for word-level CAP (TW) and phrasal-level CAP (TP). Sub-figures (a)-(c) illustrate the CAP effect on the original (Org) models, while sub-figures (d)-(f) show its impact on the fine-tuned (FT) models. Fine-tuning consistently improves performance, particularly in the middle to late layers (25\%-100\%), while early layers (0\%-25\%) show more variability and lower accuracy across models.}
    \label{fig:avg_group_acc_complete}
\end{figure*}
\section{Evaluating Parsing Accuracy and Addressing the Impact of Benepar Parser Errors}\label{sec:benepar_eval}
A key potential bias in our results comes from the reliance on the constituency parser for token-to-phrase experiments. Inaccuracies in parsing may distort the results of CAP. To address this, we report the chosen parser's accuracy by testing it on the Stanford Sentiment Treebank (SST) dataset, a dataset that offers golden labels for parsing. We aim to alleviate concerns about the parser's impact on our findings by showcasing its accuracy on the SST dataset. The parser evaluation was conducted as follows:
\paragraph{Dataset.} A subset of 1,000 randomly sampled sentences from the test split of the SST dataset was used for the analysis. The Stanford Sentiment Treebank (SST) provides annotated constituency labels, which serve as the golden labels for comparison with parser outputs. While WordNet definitions offer rich semantic information, they lack annotated golden constituency labels, making direct parser validation infeasible. The use of SST's annotations enables reliable parser evaluation and indirectly supports the validation of the parsing correctness for WordNet definitions, provided they follow standard syntactic structures.
\paragraph{Parser.} The Benepar parser was employed for parsing sentences due to its strong performance in constituency parsing tasks. Benepar is widely recognised for its robustness and ability to handle diverse syntactic structures. For this evaluation, the constituency structures generated by Benepar were directly compared against SST's golden annotations to assess its parsing accuracy.
\paragraph{Evaluation metrics.} The parser's performance was evaluated using the following metrics: (i) Precision: Proportion of correctly predicted constituents out of all predicted constituents; (ii) Recall: Proportion of correctly predicted constituents out of all ground truth constituents; (iii) F1-Score: Harmonic mean of precision and recall, providing an overall performance measure; and (iv) Accuracy: Percentage of sentences where the predicted constituency structure fully matches the ground truth.

\paragraph{Results robustness.} To ensure robustness and consistency, the evaluation was repeated across five different random seeds. This allowed for an assessment of variability in performance across multiple subsets of the dataset. Additionally, constituents were evaluated at hierarchical levels—such as root level, phrase level, and token level—to analyse parsing performance across varying syntactic granularities.

\paragraph{Results.} The evaluation yielded the following averaged metrics across five seeds for the default level of parsing (Level 1, the immediate children of the root node):
\begin{table}[h!]
\centering
\small
\begin{tabular}{|l|c|}
\hline
\textbf{Metric}       & \textbf{Mean ± Std} \\ \hline
Precision             & 0.956 ± 0.001       \\ \hline
Recall                & 0.956 ± 0.001       \\ \hline
F1-Score              & 0.956 ± 0.001       \\ \hline
Accuracy              & 0.956 ± 0.001       \\ \hline
\end{tabular}
\caption{Aggregated evaluation metrics for Level 1 constituents using the Benepar parser, averaged across five seeds.}
\label{tab:parser_eval}
\end{table}

\paragraph{Interpretation.} The results demonstrate consistently high parsing accuracy across all evaluation metrics, with minimal variability (as indicated by the low standard deviation). These findings validate the Benepar parser's reliability for parsing Level 1 constituents, which form the backbone of sentence structure. Consequently, the parser's impact on CAP results is minimal, ensuring robustness and validity of our conclusions.

\section{Detailed Performance Evaluation and Results}
\subsection{Baseline Performance}
\label{sec:baseline_perf}
Table \ref{tab:baseline_performance} summarises the baseline performance of the models used in this paper on the three tasks. The results include the accuracy of each model's original and FT versions on the test set described in Table \ref{tab:test_set_sizes}. Fine-tuning generally improves performance, particularly in the larger models such as Gemma-2B and Llama3-8B, which show notable increases in accuracy in most tasks, except the IDM task.

\begin{table*}
\centering
\small
\begin{tabular}{|p{4cm}|p{1.3cm}|p{1.5cm}|p{1.5cm}|p{1.4cm}|p{4.2cm}|}
\hline
\textbf{Task / Input Prompt} & \textbf{Model} & \textbf{CAP Layer (Type)} & \textbf{Prediction (No CAP)} & \textbf{Prediction (W/ CAP)} & \textbf{Observation / Interpretation} \\
\hline
IDM: \textit{lacking embellishment or ornamentation is called a: "} & Qwen2.5-1.5B & Layer 8 (TW)  & \textit{plain} & \textit{ornament} & Prediction shifts from correct to antonymic, likely due to token merging altering polarity. \\
\hline
IDM: \textit{remaining after all deductions is called a: "} & LLaMA3.1-8B & Layer 4 (TW)  & \textit{net} & \textit{gain} & Subtle financial distinction lost; CAP causes confusion between output and intermediate step. \\
\hline

IDM: \textit{make an effort or attempt is called a:"} & Gemma-2B & Layer 1 (TP)  & \textit{try} & \textit{<h1>} & Invalid token generation suggests breakdown in early compositional encoding. \\
\hline

IDM: \textit{a formal series of statements showing that if one thing is true something else necessarily follows from it is called a:"} & GPT2-L & Layer 24 (TP)  & \textit{proof} & \textit{form} &  Loss of logical structure leads to a more abstract or vague concept. \\
\hline

SP: \textit{"journal" is a synonym of} & Qwen2.5-1.5B & Layer 18 (TW)  & \textit{diary} & \textit{di} & Output truncated, likely due to disruption in longer multi-token word embedding. \\
\hline
SP: \textit{"get" is a synonym of} & Qwen2.5-0.5B & Layer 16 (TW)  & \textit{catch} & \textit{break} & Semantic drift under CAP; verb meaning shifts from acquisition to interruption. \\
\hline
HP: \textit{"voice" is a type of} & Gemma1-2B & Layer 16 (TW) & \textit{sound} & \textit{noise} & Precision reduced; CAP causes substitution with a noisier, less neutral concept. \\
\hline
HP: \textit{"guama" is a type of} & LLaMA3.2-3B & Layer 12 (TW) & \textit{tree} & \textit{street} & The output reflects a contextual domain shift, likely due to token-level confusion post-CAP. \\
\hline
\end{tabular}
\caption{Representative examples of model predictions with and without CAP applied at various layers. Examples highlight semantic degradation and conceptual drift caused by TW-CAP or TP-CAP applied to original models.}
\label{tab:qualitative_examples_base}
\end{table*}

\begin{table*}
\centering
\small
\begin{tabular}{|p{4cm}|p{1.3cm}|p{1.5cm}|p{1.5cm}|p{1.4cm}|p{4.2cm}|}
\hline
\textbf{Task / Input Prompt} & \textbf{Model} & \textbf{CAP Layer (Type)} & \textbf{Prediction (No CAP)} & \textbf{Prediction (W/ CAP)} & \textbf{Observation / Interpretation} \\
\hline
IDM: \textit{prepare for eating by applying heat is called a: "} & GPT2-S & Layer 4 (TW)  & \textit{cook} & \textit{heat} & CAP leads to a shift from action to cause, indicating surface-level generalisation. \\
\hline
IDM: \textit{fail to attend an event or activity is called a: "} & LLaMA3.2-3B & Layer 1 (TW) & \textit{miss} & \textit{catch} & CAP appears to invert the meaning, suggesting confusion in early compositional buildup. \\
\hline
IDM: \textit{general term for any insect or similar creeping or crawling invertebrate is called a:"} & Gemma-2B & Layer 11 (TP)  & \textit{bug} & \textit{un} & Invalid token generation suggests breakdown in compositional encoding \\
\hline
IDM: \textit{an institution of higher education created to educate and grant degrees often a part of a university is called a:"} & GPT2-S & Layer 1 (TP)  & \textit{college} & \textit{regular} & CAP reduces specificity, misclassifying to a generic adjective. \\
\hline

\hline
SP: \textit{"one fourth" is a synonym of} & Gemma1-2B & Layer 10 (TW)  & \textit{fourth} & \textit{half} & CAP merges related quantities but loses precision, leading to broader but incorrect substitution. \\
\hline
HP: \textit{"hotel" is a type of} & Qwen2.5-3B & Layer 16 (TW) & \textit{building} & \textit{room} & Shift from category to subcomponent suggests CAP disrupted higher-level abstraction. \\
\hline
HP: \textit{"hexagon" is a type of} & Qwen2.5-3B & Layer 16 (TW)  & \textit{polygon} & \textit{plane} & Hierarchical class (shape) replaced by domain (geometry); abstraction misaligned. \\
\hline
\end{tabular}
\caption{Representative examples of model predictions with and without CAP applied at various layers. Each example shows the prompt, model, CAP configuration (layer and type), predictions, and qualitative interpretation. All examples applied to fine-tuned (FT) models.}
\label{tab:qualitative_examples_ft}
\end{table*}

\subsection{Qualitative Analysis of CAP-Induced Prediction Shifts.}
Tables~\ref{tab:qualitative_examples_base} and \ref{tab:qualitative_examples_ft} present representative examples of predictions from multiple models across all the tasks, before and after CAP is applied. These examples are drawn from inputs that the model originally predicted correctly, allowing us to isolate the effects of CAP perturbations without confounding them with unrelated model failures.
Each example specifies the CAP layer, CAP type (token-to-word or token-to-phrase), and the model involved. Table~\ref{tab:qualitative_examples_base} focuses on predictions made by original (non-fine-tuned) models, while Table~\ref{tab:qualitative_examples_ft} includes outputs from fine-tuned variants. Observed shifts include truncation of multi-token terms (e.g., ``diary'' $\rightarrow$ ``di''), polarity inversion (e.g., ``plain'' $\rightarrow$ ``ornament''), loss of abstraction (``polygon'' $\rightarrow$ ``plane''), and domain misalignment (e.g., ``tree'' $\rightarrow$ ``street'').

These qualitative differences provide interpretability insights that complement the aggregate metrics reported earlier. They reveal how CAP affects not only performance but the nature of model outputs, especially in terms of semantic generalisation, abstraction shifts, and lexical precision. While we do not observe a uniform trend across layers or model families, TP-CAP consistently induces more severe semantic degradation. This suggests that as model capacity increases, internal representations may become more sensitive to disruptions from externally imposed syntactic structures, potentially due, as argued in the main paper, to a stronger reliance on learned token-level dependencies that diverge from higher-level compositional groupings. This analysis highlights the nature of semantic and lexical shifts induced by CAP, reinforcing the need for future task-specific fine-tuning strategies that improve robustness to structured representation pooling.

\subsection{Comprehensive CAP Results for All Models and Tasks}
\label{sec:full_results}
Figure~\ref{fig:avg_group_acc_complete} and Tables~\ref{tab:full_results_tw_IDM}- \ref{tab:full_results_tp_SP}, and \ref{tab:full_results_tp_HP} present the reduction in accuracy when applying word-level and phrasal CAP, respectively, across models and the three tasks: IDM, SP, and HP. The results of phrasal-level CAP for Gemma-2B and Llama3-8B are not reported due to the severe degradation in model performance under these conditions, rendering the outputs effectively unusable.

Let $A_o$ represent the original accuracy and $A_c$ represent the accuracy after applying CAP. The reported drop in accuracy, $\Delta A$, is calculated as:
\begin{equation}
    \Delta A = A_o - A_c
\end{equation}

This $\Delta A$ value is expressed in percentage points. For example, $\Delta A = 40$ indicates that the model's accuracy has decreased by 40 percentage points from its original performance, which could represent a change from $A_o = 100\%$ to $A_c = 60\%$, or any other pair of accuracies with a 40 percentage point difference. The tables report $\Delta A$ for different layer positions (1\%, 25\%, 75\%, and 100\%) in both Original and Fine-tuned settings, using three CAP protocols: Max, Mean, and Sum. This representation allows for a direct comparison of CAP's impact across different models and tasks, independent of their baseline performance levels.

 %with the best (green) and worst (red) values highlighted to indicate performance changes.

\begin{table*}[]
\centering
\small
\begin{tabular}{|l|c|ccc|ccc|}
\hline
\multirow{2}{*}{\textbf{Model}} & \multirow{2}{*}{\textbf{Layer Position}} & \multicolumn{3}{c|}{\textbf{Original}} & \multicolumn{3}{c|}{\textbf{Fine-tuned}} \\
\cline{3-8}
&  & \textbf{Max} & \textbf{Mean} & \textbf{Sum} & \textbf{Max} & \textbf{Mean} & \textbf{Sum} \\
\hline
\multicolumn{8}{|c|}{\textbf{IDM (Inverse Dictionary Modelling)}} \\
\hline
\multirow{4}{*}{\textbf{GPT2-small}} 
& 1\%  & 4.76\%  & 4.44\% & 4.69\%  & 8.04\% & 7.72\% & 7.22\% \\
& 25\% & 3.09\%  & 2.74\% & 3.26\%  & 5.87\% & 5.85\% & 6.24\% \\
& 75\% & 2.64\%  & 2.36\% & 2.74\%  & 2.72\% & 2.47\% & 2.35\% \\
& 100\%& 1.43\%  & 1.24\%  & 1.24\%   & 0.46\% & 0.39\% & 0.39\% \\
\hline
\multirow{4}{*}{\textbf{GPT2-medium}} 
& 1\%  & 16.75\%  & 16.36\% & 13.77\%    & 24.51\%  & 12.70\%  & 7.44\% \\
& 25\% & 6.73\%   & 5.692\% & 6.22\%     & 5.04\%   & 4.84\% & 5.36\% \\
& 75\% & 18.61\%  & 2.13\%  & 2.89\%     & 11.79\%  & 2.09\% & 1.72\% \\
& 100\%& 1.58\%   & 0.41\%  & 0.41\%     & 2.27\%   & 1.29\% & 1.29\%  \\
\hline
\multirow{4}{*}{\textbf{GPT2-large}} 
& 1\%  & 8.06\%   & 9.15\%  & 6.70\% & 10.61\% & 10.01\% & 7.83\% \\
& 25\% & 5.19\%  & 4.94\% &  5.63\%  & 6.25\% & 5.77\% & 6.32\% \\
& 75\% & 5.28\%  & 2.62\% &  2.39\%   & 3.66\% & 1.62\% & 0.88\% \\
& 100\%& 0.84\%  & 0.12\%  & 0.19\%   & 0.22\% & 0.16\% & 0.16\% \\
\hline
\multirow{4}{*}{\textbf{Gemma-2B}} 
& 1\%  & 97.91\%   & 23.51\%  & 23.75\%  & 57.58\% & 22.70\% & 21.99\% \\
& 25\% & 86.32\%   & 16.20\% & 19.27\%  & 50.45\%  & 14.08\% & 15.57\% \\
& 75\% & 52.38\%   & 31.03\% & 24.74\% & 21.77\% & 14.99\% & 12.80\% \\
& 100\%& 6.87\%   & 10.61\% & 10.61\% & 2.21\% & 2.05\% & 2.05\% \\
\hline
\multirow{4}{*}{\textbf{Llama3-8B}} 
& 1\%  & 25.49\%   & 24.99\%  & 24.94\%  & 24.44\% & 23.42\% & 23.48\% \\
& 25\% & 20.02\%   & 5.87\% & 5.74\%  & 8.81\%  & 6.03\% & 5.92\% \\
& 75\% & 7.31\%   & 3.40\% & 3.54\% & 5.16\% & 3.47\% & 3.29\% \\
& 100\%& 2.80\%   & 1.77\% & 1.77\% & 1.55\% & 1.33\% & 1.33\% \\
\hline

\multirow{4}{*}{\textbf{Llama3-3B}} 
& 1\%  & 28.79\%  & 26.36\% & 25.96\%& 25.54\% & 22.71\% & 22.74\% \\
& 25\% & 31.73\%  & 8.08\% & 6.99\%  & 13.44\%  & 5.84\% & 5.8\% \\
& 75\% & 12.27\%  & 5.84\% & 5.22\%  & 8.54\% & 5.03\% & 5.15\% \\
& 100\%& 3.62\%   & 1.99\% & 1.99\%  & 2.37\% & 1.82\% & 1.85\% \\
\hline

% \multirow{4}{*}{\textbf{Llama3-8B Instruct}} 
% & 1\%  & 41.75\% & 48.99\% & 45.99\%& 22.01\% & 19.97\% & 20.29\% \\
% & 25\% & 44.43\% & 25.22\% & 25.12\%& 17.62\% & 13.51\% & 15.22\% \\
% & 75\% & 26.17\% & 24.42\% & 24.98\%& 7.9\% & 5.34\% & 5.19\% \\
% & 100\%& 21.8\%  & 12.04\% & 12.04\%& 1.51\% & 1.29\% & 1.33\% \\
% \hline
\multirow{4}{*}{\textbf{Qwen2.5-0.5B}} 
& 1\%  & 10.12\% & 8.2\% & 8.23\%& 7.85\% & 6.39\% & 6.00\% \\
& 25\% & 5.19\%  & 4.21\% & 4.45\%& 4.35\% & 3.29\% & 3.49\% \\
& 75\% & 3.56\%  & 2.82\% & 3.15\%& 2.39\% & 2.24\% & 2.15\% \\
& 100\%& 0.98\%  & 0.98\% & 0.98\%& 0.23\% & 0.28\% & 0.33\% \\
\hline
\multirow{4}{*}{\textbf{Qwen2.5-1.5B}} 
& 1\%  & 14.56\%  & 11.04\% & 10.22\%& 9.47\% & 7.36\% & 7.48\% \\
& 25\% & 13.29\%  & 4.45\% & 5.34\%& 6.83\% & 3.86\% & 4.00\% \\
& 75\% & 7.03\%  & 2.68\% & 2.84\%& 4.21\% & 2.74\% & 2.79\% \\
& 100\%& 0.7\%  & 0.4\% & 0.4\%& 0.65\% & 0.23\% & 0.23\% \\
\hline
\multirow{4}{*}{\textbf{Qwen2.5-3B}} 
& 1\%  & 12.63\% & 12.27\%& 11.44\%& 7.85\% & 6.71\% & 6.48\% \\
& 25\% & 18.61\% & 8.59\% & 9.11\%& 10.66\% & 4.75\% & 5.82\% \\
& 75\% & 7.23\%  & 4.00\% & 3.79\%& 3.65\% & 2.83\% & 2.8\% \\
& 100\%& 0.39\%  & 0.4\% & 0.4\%& 0.31\% & 0.17\% & 0.2\% \\
\hline
\end{tabular}
\caption{Performance drop (in percentage points) for GPT2 (small, medium, large), Gemma-2B, Llama3 (3B, 8B), and Qwen2.5 (0.5B, 1.5B, 3B) models after applying word-level CAP for the Inverse Dictionary Modelling (IDM) task. Results are reported for different layer positions (1\%, 25\%, 75\%, and 100\%) in both Original and Fine-tuned settings, using three CAP protocols: Max, Mean, and Sum.}
\label{tab:full_results_tw_IDM}
\end{table*}

\begin{table*}[ht]
\centering
\small
\begin{tabular}{|l|c|ccc|ccc|}
\hline
\multirow{2}{*}{\textbf{Model}} & \multirow{2}{*}{\textbf{Layer Position}} & \multicolumn{3}{c|}{\textbf{Original}} & \multicolumn{3}{c|}{\textbf{Fine-tuned}} \\
\cline{3-8}
&  & \textbf{Max} & \textbf{Mean} & \textbf{Sum} & \textbf{Max} & \textbf{Mean} & \textbf{Sum} \\
\hline
\multicolumn{8}{|c|}{\textbf{SP (Synonym Prediction)}} \\
\hline
\multirow{4}{*}{\textbf{GPT2-small}} 
& 1\%  & 99.04\%   & 99.04\%  & 99.04\%  & 59.68\% & 49.40\% & 34.68\% \\
& 25\% & 98.56\%    & 98.56\% & 97.60\%  & 61.09\% & 30.85\% & 29.64\% \\
& 75\% & 96.15\%    & 94.23\% & 93.75\%   & 40.12\%  & 9.68\% & 10.48\% \\
& 100\%& 6.73\%     & 7.21\%  & 7.21\%   & 3.23\%   & 2.42\% & 2.42\% \\
\hline
\multirow{4}{*}{\textbf{GPT2-medium}} 
& 1\%  & 96.43\%   & 96.43\% & 96.43\%  & 83.35\%   & 82.50\%  & 84.06\% \\
& 25\% & 96.13\%   & 96.43\% & 96.43\%  & 79.22\%   & 80.22\% & 80.79\%\\
& 75\% & 63.93\%   & 48.30\% & 56.63\% & 48.36\%   & 23.23\% & 24.53\%  \\
& 100\%& 6.68\%  & 3.41\% & 3.41\% & 6.55\%   & 5.12\% & 5.12\%  \\
\hline
\multirow{4}{*}{\textbf{GPT2-large}} 
& 1\%  & 98.49\%   & 98.49\%  & 98.06\%  & 78.61\% & 78.33\% & 80.17\% \\
& 25\% & 97.63\%   & 97.63\% & 97.63\%  & 80.93\%  & 81.78\% & 79.89\% \\
& 75\% & 34.27\%   & 27.59\% & 28.52\% & 11.91\% & 10.02\% & 10.49\% \\
& 100\%& 1.29\%   & 1.51\% & 1.51\% & 1.22\% & 39.12\% & 0.61\% \\
\hline
\multirow{4}{*}{\textbf{Gemma-2B}} 
& 1\%  & 99.99\%    & 99.80\%   & 83.47\%   & 99.93\%  & 99.15\%  & 96.38\%  \\
& 25\% & 99.99\%    & 97.46\%  & 63.68\%   & 90.20\%     & 90.24\%  & 65.82\%  \\
& 75\% & 84.63\%  & 60.66\%  & 61.15\%  & 89.87\%  & 75.68\%  & 68.65\%  \\
& 100\%& 4.30\%    & 8.69\%  & 8.69\%  & 2.98\%  & 4.57\%  & 4.57\% \\

\hline
\multirow{4}{*}{\textbf{Llama3-8B}} 
& 1\%  & 99.99\%   & 99.90\%  & 99.90\%  & 99.99\% & 99.88\% & 99.88\% \\
& 25\% & 85.55\%   & 83.50\% & 82.81\%  & 87.63\%   & 85.75\% & 85.63\% \\
& 75\% & 53.35\%   & 50.55\% & 49.77\% & 31.29\% & 30.29\% & 29.91\% \\
& 100\%& 9.28\%   & 9.96\% & 9.96\% & 5.20\% & 5.82\% & 5.82\% \\

\hline

\multirow{4}{*}{\textbf{Llama3-3B}} 
& 1\%  & 100\%   & 100\%  & 100\%  & 100\% & 100\% & 100\% \\
& 25\% & 85.81\%   & 86.2\% & 85.16\%  & 88.47\%   & 84.54\% & 85.48\% \\
& 75\% &40.18\%   & 39.3\% & 38.91\% & 14.77\% & 16.48\% & 15.64\% \\
& 100\%& 5.77\%   & 6.16\% & 6.16\% & 5.8\% & 6.12\% & 6.12\% \\

\hline

% \multirow{4}{*}{\textbf{Llama3-8B (Instruct)}} 
% & 1\%  & 85.46\%   & 85.71\%  & 83.88\%  & 42.9\% & 32.33\% & 31.09\% \\
% & 25\% & 54.34\%   & 55.48\% & 58.83\%  & 27.52\%   & 18.13\% & 18.27\% \\
% & 75\% & 44.92\%   & 31.51\% & 25.77\% & 19.23\% & 12.5\% & 12.23\% \\
% & 100\%& 6.81\%   & 7.3\% & 7.3\% & 5.77\% & 4.81\% & 4.95\% \\
% \hline

\multirow{4}{*}{\textbf{Qwen2.5-0.5B}} 
& 1\%  & 81.77\%   & 88.89\%  & 79.17\%  & 64.24\% & 58.36\% & 53.3\% \\
& 25\% & 90.8\%   & 91.15\% & 86.11\%  & 54.51\%   & 54.38\% & 37.22\% \\
& 75\% & 63.72\%   & 66.32\% & 39.06\% & 48.87\% & 48.57\% & 24.29\% \\
& 100\%& 8.51\%   & 10.07\% & 8.51\% & 3.67\% & 3.8\% & 3.8\% \\
\hline

\multirow{4}{*}{\textbf{Qwen2.5-1.5B}} 
& 1\%  & 89.35\%   & 84.52\%  & 84.23\%  & 64.55\% & 56.79\% & 56.03\% \\
& 25\% & 90.58\%   & 83.48\% & 83.19\%  & 60.45\%   & 55.5\% & 54.79\% \\
& 75\% & 22.06\%   & 22.21\% & 18.8\% & 10.88\% & 10.34\% & 10.02\% \\
& 100\%& 6.82\%   & 3.55\% & 3.55\% & 8.19\% & 7.87\% & 7.87\% \\
\hline

\multirow{4}{*}{\textbf{Qwen2.5-3B}} 
& 1\%  & 81.39\% & 81.53\%  & 73.58\%  & 55.93\% & 49.35\% & 49.57\% \\
& 25\% & 93.04\% & 89.91\% & 82.81\%  & 72.41\%   & 42.78\% & 38.47\% \\
& 75\% & 77.84\% & 69.6\% & 49.43\% & 43.24\% & 22.13\% & 15.25\% \\
& 100\%& 3.98\%  & 3.13\% & 3.13\% & 1.4\% & 1.29\% & 1.29\% \\
\hline

\end{tabular}
\caption{Performance drop (in percentage points) for GPT2 (small, medium, large), Gemma-2B, Llama3 (3B, 8B), and Qwen2.5 (0.5B, 1.5B, 3B) models after applying word-level CAP for the Synonym Prediction (SP) task. Results are reported for different layer positions (1\%, 25\%, 75\%, and 100\%) in both Original and Fine-tuned settings, using three CAP protocols: Max, Mean, and Sum.}
\label{tab:full_results_tw_SP}
\end{table*}

\begin{table*}[ht]
\centering
\small
\begin{tabular}{|l|c|ccc|ccc|}
\hline
\multirow{2}{*}{\textbf{Model}} & \multirow{2}{*}{\textbf{Layer Position}} & \multicolumn{3}{c|}{\textbf{Original}} & \multicolumn{3}{c|}{\textbf{Fine-tuned}} \\
\cline{3-8}
&  & \textbf{Max} & \textbf{Mean} & \textbf{Sum} & \textbf{Max} & \textbf{Mean} & \textbf{Sum} \\
\hline

\multicolumn{8}{|c|}{\textbf{HP (Hypernym Prediction)}} \\
\hline
\multirow{4}{*}{\textbf{GPT2-small}} 
& 1\%  & 99.75\%   & 99.75\%  & 99.75\%  & 91.19\%   & 91.08\%  & 88.20\% \\
& 25\% & 99.47\%   & 99.29\% & 98.94\%  & 81.35\%   & 76.76\% & 72.63\% \\
& 75\% & 95.40\% & 91.16\% & 91.32\%   & 48.75\% & 38.54\% & 38.40\%  \\
& 100\%& 8.12\% & 6.39\% & 6.39\%   & 1.35\% & 1.38\% & 1.28\% \\
\hline
\multirow{4}{*}{\textbf{GPT2-medium}} 
& 1\%  & 99.42\%   & 99.40\%  & 99.44\%  & 93.42\%  & 92.17\%  & 91.69\% \\
& 25\% & 99.11\%   & 98.55\% & 97.85\%  &  91.64\% & 86.11\%  & 85.76\% \\
& 75\% & 74.83\%   & 33.22\% & 41.52\%  & 3.86\%   & 2.23\% & 2.33\% \\
& 100\%& 4.42\%   & 1.79\% & 1.79\%     & 3.86\% & 2.23\%  & 2.32\%  \\
\hline

\multirow{4}{*}{\textbf{GPT2-large}} 
& 1\%  & 99.27\%   & 99.32\%  & 99.20\%  & 91.49\% & 90.90\% & 89.80\% \\
& 25\% & 98.81\%   & 98.75\% & 98.10\%  & 87.30\%   & 87.54\% & 84.16\% \\
& 75\% & 45.17\%   & 29.85\% & 35.66\% & 7.61\% & 6.89\% & 6.22\% \\
& 100\%& 2.14\%   & 0.45\% & 0.90\% & 0.69\% & 0.50\% & 0.56\% \\
\hline

\multirow{4}{*}{\textbf{Gemma-2B}} 
& 1\%  & 99.99\%   & 98.97\%  & 70.22\%  & 99.88\% & 95.39\% & 74.03\% \\
& 25\% & 99.98\%   & 90.58\% & 86.35\%  & 90.98\%   & 73.78\% & 86.01\% \\
& 75\% & 68.14\%   & 80.06\% & 80.20\% & 58.56\% & 72.57\% & 66.56\% \\
& 100\%& 5.89\%   & 10.99\% & 10.99\% & 1.58\% & 2.12\% & 2.12\% \\
\hline
\multirow{4}{*}{\textbf{Llama3-8B}} 
& 1\%  & 99.99\%    & 99.99\%   & 99.14\%   & 99.99\%    & 99.10\%   & 99.14\%  \\
& 25\% & 80.85\%    & 76.97\%  & 76.81\%   & 72.67\%   & 71.86\% & 71.40\% \\
& 75\% & 24.43\% & 24.39\% & 23.11\%    & 19.65\%  & 19.71\%  & 18.77\%   \\
& 100\%& 3.83\%  & 4.49\%  & 4.49\%    & 4.63\%  & 4.04\%  & 4.20\%  \\
\hline
\multirow{4}{*}{\textbf{Llama3-3B}} 
& 1\%  & 100\%    & 99.95\%   & 99.95\%   & 99.93\%    & 99.86\%   & 99.82\%  \\
& 25\% & 88.04\%    & 83.87\%   & 84.34\%   & 65.53\%    & 63.92\%   & 64.17\%  \\
& 75\% & 26.06\%    & 24.47\%   & 23.4\%   & 11.06\%    & 10.52\%   & 10.79\%  \\
& 100\%& 4.34\%    & 4.31\%   & 4.31\%   & 3.85\%    & 4.08\%   & 3.86\%  \\
\hline

% \multirow{4}{*}{\textbf{Llama3-8B (Instruct)}} 
% & 1\%  & 92.51\%    & 92.45\%   & 92.48\%   & 95.86\%    & 96.21\%   & 96.03\%  \\
% & 25\% & 68.41\%    & 65.9\%   & 67.24\%   & 70.72\%    & 69.99\%   & 69.64\%  \\
% & 75\% & 15.79\%    & 14.78\%   & 15.55\%   & 20\%    & 20.29\%   & 19.53\%  \\
% & 100\%& 0.57\%    & 0.39\%   & 0.39\%   & 3.57\%    & 3.5\%   & 3.29\%  \\
% \hline

\multirow{4}{*}{\textbf{Qwen2.5-0.5B}} 
& 1\%  & 93.76\%    & 90.95\%   & 85.27\%   & 86.33\%    & 80.55\%   & 77.91\%  \\
& 25\% & 97.12\%    & 97.51\%   & 89.18\%   & 74.83\%    & 75.41\%   & 75.77\%  \\
& 75\% & 76.74\%    & 77.96\%   & 55.39\%   & 50.69\%    & 49.71\%   & 48.81\%  \\
& 100\%& 6.15\%    & 5.56\%   & 5.56\%   & 2.48\%    & 2.34\%   & 2.34\%  \\
\hline

\multirow{4}{*}{\textbf{Qwen2.5-1.5B}} 
& 1\%  & 97.14\%    & 90.5\%   & 88.96\%   & 88.52\%    & 83.19\%   & 77.21\%  \\
& 25\% & 98.12\%    & 95.66\%   & 94.04\%   & 72.29\%    & 68.18\%   & 68.33\%  \\
& 75\% & 18.27\%    & 18.72\%   & 17.94\%   & 8.94\%    & 9.64\%   & 9.51\%  \\
& 100\%& 7.13\%    & 6.81\%   & 6.81\%   & 3.95\%    & 3.8\%   & 3.8\%  \\
\hline
\multirow{4}{*}{\textbf{Qwen2.5-3B}} 
& 1\%  & 83.26\%    & 82.41\%   & 68.8\%   & 75.13\%    & 72.56\%   & 70.69\%  \\
& 25\% & 97.36\%    & 96.32\%   & 88.81\%   & 92.69\%    & 79.67\%   & 79.63\%  \\
& 75\% & 86.56\%    & 71.45\%   & 45.47\%   & 40.87\%    & 30.95\%   & 33.04\%  \\
& 100\%& 2.07\%    & 1.89\%   & 1.89\%   & 0.45\%    & 0.35\%   & 0.41\%  \\
\hline

\end{tabular}
\caption{Performance drop (in percentage points) for GPT2 (small, medium, large), Gemma-2B, Llama3 (3B, 8B), and Qwen2.5 (0.5B, 1.5B, 3B) models after applying word-level CAP for the Hypernym Prediction (HP) task. Results are reported for different layer positions (1\%, 25\%, 75\%, and 100\%) in both Original and Fine-tuned settings, using three CAP protocols: Max, Mean, and Sum.}
\label{tab:full_results_tw_HP}
\end{table*}

\begin{table*}[ht]
\centering
\small
\begin{tabular}{|l|c|ccc|ccc|}
\hline
\multirow{2}{*}{\textbf{Model}} & \multirow{2}{*}{\textbf{Layer Position}} & \multicolumn{3}{c|}{\textbf{Original}} & \multicolumn{3}{c|}{\textbf{Fine-tuned}} \\
\cline{3-8}
&  & \textbf{Max} & \textbf{Mean} & \textbf{Sum} & \textbf{Max} & \textbf{Mean} & \textbf{Sum} \\
\hline
\multicolumn{8}{|c|}{\textbf{IDM (Inverse Dictionary Modelling)}} \\
\hline
\multirow{4}{*}{\textbf{GPT2-small}} 
& 1\%  & 93.00\%  & 93.94\% & 96.56\%  & 77.912\% & 77.73\% & 80.28\% \\
& 25\% & 90.20\%  & 87.85\% & 91.41\%  & 65.73\% & 62.95\% & 72.31\% \\
& 75\% & 87.81\%  & 78.66\% & 84.90\%  & 55.74\% & 46.81\% & 55.73\% \\
& 100\%& 48.10\%  & 45.10\% & 38.04\%  & 11.11\% & 8.45\% & 8.11\% \\
\hline
\multirow{4}{*}{\textbf{GPT2-medium}} 
& 1\%  & 87.96\%  & 89.87\% & 92.52\%    & 81.12\%  & 82.37\%  & 81.83\% \\
& 25\% & 77.06\%   & 82.71\% & 86.54\%     & 69.53\%   & 75.19\% & 77.55\% \\
& 75\% & 76.35\%  & 48.76\%  & 57.68\%     & 60.60\%  & 29.52\% & 33.12\% \\
& 100\%& 29.23\%   & 23.12\%  & 23.21\%     & 13.03\%   & 9.75\% & 9.94\%  \\
\hline
\multirow{4}{*}{\textbf{GPT2-large}} 
& 1\%  & 87.06\%   & 89.91\%  & 88.44\% & 81.14\% & 85.35\% & 79.46\% \\
& 25\% & 73.54\%  & 78.18\% &  82.48\%  & 69.39\% & 73.85\% & 71.90\% \\
& 75\% & 49.02\%  & 42.06\% &  40.38\%   & 20.59\% & 19.78\% & 21.45\% \\
& 100\%& 28.14\%  & 24.22\%  & 24.78\%   & 6.46\% & 6.67\% & 8.44\% \\
\hline
\multirow{4}{*}{\textbf{Qwen2.5-0.5B}} 
& 1\%  & 93.97\%  & 91.19\% & 87.15\%  & 90.94\%  & 84.44\%  & 78.85\%  \\
& 25\% & 84.64\%  & 76.78\% &78.00\%  & 76.36\%  & 66.24\%  & 67.16\% \\
& 75\% & 61.75\%  & 57.95\% & 63.86\%  & 48.86\%  & 41.8\%  & 46.25\%   \\
& 100\%& 32.29\%  &26.8\% & 19.5\%  & 13.55\%  &10.17\%  & 15.08\%  \\
\hline
\multirow{4}{*}{\textbf{Qwen2.5-1.5B}} 
& 1\%  & 98.24\% &  95.8\% & 95.82\%  & 93.31\%  & 87.33\%  & 80.81\%  \\
& 25\% & 96.4\%  & 84.72\% &89.41\%  & 79.52\%  & 63.00\%  & 65.53\% \\
& 75\% & 69.68\%  & 64.6\% & 60.33\%  & 19.11\%  & 14.72\%  & 24.01\%   \\
& 100\%& 68.03\%  &60.04\% & 56.6\%  & 12.01\%  &7.46\%  & 12.72\%  \\
\hline

\multirow{4}{*}{\textbf{Qwen2.5-3B}} 
& 1\%  & 96.51\%  & 94.37\% & 94.64\%  & 90.11\%  & 86.02\%  & 80.57\%  \\
& 25\% & 96.82\%  & 89.89\% &92.39\%  & 90.24\%  & 76.55\%  & 76.28\% \\
& 75\% & 82.27\%  & 74.71\% & 77.07\%  & 47.45\%  & 36.06\%  & 39.95\%   \\
& 100\%& 62.26\%  &62.21\% & 58.12\%  & 7.41\%  &5.52\%  & 8.18\%  \\
\hline

% \multirow{4}{*}{\textbf{Llama3-8B}} 
% & 1\%  & 99.99\%   & 99.99\%  & 99.99\%  & 99.99\% & 99.99\% & 99.99\% \\
% & 25\% & 99.99\%   & 99.99\% & 99.99\%  & 99.99\%  & 99.99\% & 99.99\% \\
% & 75\% & 99.99\%   & 99.99\% & 99.99\% & 99.99\% & 99.99\% & 99.99\% \\
% & 100\%& 99.99\%   & 99.99\% & 99.99\% & 99.99\% & 99.99\% & 99.99\% \\
% \hline

\end{tabular}
\caption{Performance drop (in percentage points) for GPT2-small, GPT2-medium, and GPT2-large models after applying phrasal-level CAP across three tasks: Inverse Dictionary Modelling (IDM), Synonym Prediction (SP), and Hypernym Prediction (HP). Results are reported for different layer positions (1\%, 25\%, 75\%, and 100\%) in both Original and Fine-tuned settings, using three CAP protocols: Max, Mean, and Sum. Results for Gemma-2B and Llama3-8B are omitted due to severe performance degradation under phrasal-level CAP.}
\label{tab:full_results_tp_IDM}
\end{table*}

\begin{table*}[ht]
\centering
\small
\begin{tabular}{|l|c|ccc|ccc|}
\hline
\multirow{2}{*}{\textbf{Model}} & \multirow{2}{*}{\textbf{Layer Position}} & \multicolumn{3}{c|}{\textbf{Original}} & \multicolumn{3}{c|}{\textbf{Fine-tuned}} \\
\cline{3-8}
&  & \textbf{Max} & \textbf{Mean} & \textbf{Sum} & \textbf{Max} & \textbf{Mean} & \textbf{Sum} \\
\hline

% \multirow{4}{*}{\textbf{Llama3-8B}} 
% & 1\%  & 99.99\%   & 99.99\%  & 99.99\%  & 99.99\% & 99.99\% & 99.99\% \\
% & 25\% & 99.99\%   & 99.99\% & 99.99\%  & 99.99\%  & 99.99\% & 99.99\% \\
% & 75\% & 99.99\%   & 99.99\% & 99.99\% & 99.99\% & 99.99\% & 99.99\% \\
% & 100\%& 99.99\%   & 99.99\% & 99.99\% & 99.99\% & 99.99\% & 99.99\% \\
% \hline
\multicolumn{8}{|c|}{\textbf{SP (Synonym Prediction)}} \\
\hline
\multirow{4}{*}{\textbf{GPT2-small}} 
& 1\%  & 99.99\%   & 99.99\%& 99.99\%& 64.90\% & 58.47\% & 53.22\% \\
& 25\% & 92.97\%    & 93.36\%& 93.36\%& 61.27\% & 37.19\% & 74.69\% \\
& 75\% & 92.58\%& 90.63\%& 92.19\%& 43.35\%  & 20.57\% & 52.22\% \\
& 100\%& 58.46\% & 47.92\%& 51.43\%   & 13.27\%   & 7.57\% & 12.45\% \\
\hline
\multirow{4}{*}{\textbf{GPT2-medium}} 
& 1\%  & 97.55\%   & 95.11\% & 99.99\%    & 88.92\%   & 84.23\%  & 84.80\% \\
& 25\% & 97.55\%   & 99.73\% & 97.55\%  & 75.00\%   & 76.85\% & 85.65\%\\
& 75\% & 71.20\%   & 68.21\% & 77.45\% & 47.72\%   & 22.16\% & 45.88\%  \\
& 100\%& 66.30\%  & 39.40\% & 52.17\% & 12.93\%   & 6.68\% & 9.52\%  \\
\hline
\multirow{4}{*}{\textbf{GPT2-large}} 
& 1\%  & 96.67\%   & 98.33\%  & 96.67\%  & 92.55\% & 80.76\% & 79.58\% \\
& 25\% & 96.67\%   & 96.44\% & 97.90\%  & 79.44\%  & 80.48\% & 82.86\% \\
& 75\% & 78.83\%   & 66.72\% & 66.32\% & 18.63\% & 15.80\% & 21.00\% \\
& 100\%& 67.10\%   & 45.83\% & 56.68\% & 9.69\% & 7.15\% & 8.33\% \\
\hline
\multirow{4}{*}{\textbf{Qwen2.5-0.5B}} 
& 1\%  & 99.32\%  & 95.88\% & 92.87\%  & 81.67\%  & 61.89\%  & 57.95\%  \\
& 25\% & 98.65\%  & 95.91\% & 96.45\%  & 60.19\%  & 58.75\%  & 58.43\% \\
& 75\% & 93.21\%  & 84.66\% & 77.4\%  & 56.29\%  & 49.3\%  & 44.94\%   \\
& 100\%& 68.78\%  & 45.74\% & 43.92\%  & 13.56\%  &7.47\%  & 16.79\%  \\
\hline
\multirow{4}{*}{\textbf{Qwen2.5-1.5B}} 
& 1\%  & 98.1\%  & 96.33\% & 94.43\%  & 72.33\%  & 58.5\%  & 59.55\%  \\
& 25\% & 97.55\%  & 96.2\% &95.38\%  & 63.79\%  & 55.84\%  & 68.93\% \\
& 75\% & 75.72\%  & 55.17\% & 48.41\%  & 19.33\%  &14.48\%  & 26.87\%   \\
& 100\%& 70.39\%  &38.68\% & 36.29\%  & 18.73\%  &10.41\%  & 20.97\%  \\
\hline
\multirow{4}{*}{\textbf{Qwen2.5-0.5B}} 
& 1\%  & 96.47\%  & 95.52\% & 90.31\%  & 74.05\%  & 67.1\%  & 56.57\%  \\
& 25\% & 99.32\%  & 98.1\% &94.29\%  & 94.89\%  & 56.93\%  & 57.38\% \\
& 75\% & 94.02\%  & 89.46\% & 83.4\%  & 86.43\%  & 64.01\%  & 43.39\%   \\
& 100\%& 47.00\%  &35.56\% & 31.32\%  & 20.07\%  &15.19\%  & 21.15\%  \\
\hline

% \multirow{4}{*}{\textbf{Llama3-8B}} 
% & 1\%  & 99.99\%  & 99.99\% & 99.99\%  & 99.99\%  & 99.99\%  & 99.99\%  \\
% & 25\% & 99.99\%  & 99.99\% & 99.99\%  & 99.99\%  & 99.99\%  & 99.99\% \\
% & 75\% & 99.99\%  & 99.99\% & 99.99\%  & 99.99\%  & 99.99\%  & 99.99\%   \\
% & 100\%& 99.99\%  & 99.99\% & 99.99\%  & 99.99\%  & 99.99\%  & 99.99\%  \\
% \hline

\end{tabular}
\caption{Performance drop (in percentage points) for GPT2-small, GPT2-medium, and GPT2-large models after applying phrasal-level CAP across three tasks: Inverse Dictionary Modelling (IDM), Synonym Prediction (SP), and Hypernym Prediction (HP). Results are reported for different layer positions (1\%, 25\%, 75\%, and 100\%) in both Original and Fine-tuned settings, using three CAP protocols: Max, Mean, and Sum. Results for Gemma-2B and Llama3-8B are omitted due to severe performance degradation under phrasal-level CAP.}
\label{tab:full_results_tp_SP}
\end{table*}

\begin{table*}[ht]
\centering
\small
\begin{tabular}{|l|c|ccc|ccc|}
\hline
\multirow{2}{*}{\textbf{Model}} & \multirow{2}{*}{\textbf{Layer Position}} & \multicolumn{3}{c|}{\textbf{Original}} & \multicolumn{3}{c|}{\textbf{Fine-tuned}} \\
\cline{3-8}
&  & \textbf{Max} & \textbf{Mean} & \textbf{Sum} & \textbf{Max} & \textbf{Mean} & \textbf{Sum} \\
\hline
\multicolumn{8}{|c|}{\textbf{HP (Hypernym Prediction)}} \\
\hline
\multirow{4}{*}{\textbf{GPT2-small}} 
& 1\%  & 99.40\% & 99.26\%  & 47.24\%  & 89.31\%   & 89.86\%  & 88.76\% \\
& 25\% & 99.31\% & 98.12\% & 46.38\%  & 77.72\%   & 73.12\% & 76.08\% \\
& 75\% & 95.63\% & 91.78\% & 45.57\%   & 47.73\% & 336.59\% & 48.32\%  \\
& 100\%& 65.62\% & 45.84\% & 34.80\%   & 4.80\% & 3.64\% & 4.00\% \\
\hline
\multirow{4}{*}{\textbf{GPT2-medium}} 
& 1\%  & 99.77\%   & 99.56\%  & 99.950\%   & 92.67\% & 90.40\%  & 92.54\% \\
& 25\% & 99.92\%   & 99.35\% & 99.47\%   & 90.38\% &  84.29\%  & 86.84\% \\
& 75\% & 77.77\%   & 58.17\% & 80.58\%  & 63.00\% &  21.55\% & 23.32\% \\
& 100\%&59.28\%   & 27.47\% & 30.54\%     & 8.46\%  &  5.10\%  & 5.10\%  \\
\hline

\multirow{4}{*}{\textbf{GPT2-large}} 
& 1\%  & 99.77\%  & 99.71\%  & 99.76\% & 91.63\% & 92.56\% & 88.92\% \\
& 25\% & 99.82\%  & 98.72\%  & 98.82\% & 85.31\% & 85.35\% & 84.58\% \\
& 75\% & 66.58\%  & 49.79\%  & 63.56\% & 9.87\%   & 8.79\% & 9.73\% \\
& 100\%& 35.57\%  & 24.79\%   & 26.69\%  & 6.99\%  & 5.05\% & 4.82\% \\
\hline

\multirow{4}{*}{\textbf{Qwen2.5-0.5B}} 
& 1\%  & 99.06\%  & 97.77\% & 92.97\%  & 94.46\%  & 81.39\%  & 79.64\%  \\
& 25\% & 99.85\%  & 98.54\% &96.95\%  & 75.14\%  & 76.07\%  & 86.94\% \\
& 75\% & 94.87\%  & 87.81\% & 88.37\%  & 56.27\%  & 53.09\%  & 63.33\%   \\
& 100\%& 68.71\%  &27.91\% & 27.92\%  & 10.6\%  &7.68\%  & 15.16\%  \\
\hline

\multirow{4}{*}{\textbf{Qwen2.5-1.5B}} 
& 1\%  & 99.81\%  & 97.07\% & 92.75\%  & 90.34\%  & 84.61\%  & 78.76\%  \\
& 25\% & 99.64\%  & 97.97\% &96.98\%  & 72.81\%  & 68.48\%  & 77.13\% \\
& 75\% & 84.28\%  & 47.63\% & 43.15\%  & 17.12\%  & 14.76\%  & 28.18\%   \\
& 100\%& 82.22\%  &26.00\% & 27.7\%  & 13.49\%  &9.08\%  & 17.98\%  \\
\hline

\multirow{4}{*}{\textbf{Qwen2.5-3B}} 
& 1\%  & 93.95\%  & 91.81\% & 82.05\%  & 77.6\%  & 73.86\%  & 71.41\%  \\
& 25\% & 99.24\%  & 98.54\% & 95.97\%  & 93.6\%  & 80.32\%  & 80.77\% \\
& 75\% & 94.48\%  & 88.91\% & 78.88\%  & 54.32\%  & 38.19\%  & 57.87\%   \\
& 100\%& 55.28\%  &27.4\% & 25.1\%  & 15.1\%  &8.77\%  & 13.77\%  \\
\hline

\end{tabular}
\caption{Performance drop (in percentage points) for GPT2-small, GPT2-medium, and GPT2-large models after applying phrasal-level CAP across three tasks: Inverse Dictionary Modelling (IDM), Synonym Prediction (SP), and Hypernym Prediction (HP). Results are reported for different layer positions (1\%, 25\%, 75\%, and 100\%) in both Original and Fine-tuned settings, using three CAP protocols: Max, Mean, and Sum. Results for Gemma-2B and Llama3-8B are omitted due to severe performance degradation under phrasal-level CAP.}
\label{tab:full_results_tp_HP}
\end{table*}

\end{document}